\documentclass[letterpaper, 10pt, conference]{ieeeconf}
\IEEEoverridecommandlockouts
\makeatletter
\let\NAT@parse\undefined
\makeatother
\usepackage{algorithm}
\usepackage{algorithmic}
\usepackage{amsfonts}
\usepackage{amsmath}
\usepackage{amssymb}
\usepackage{array}
\usepackage{bm}
\usepackage{breakcites}
\usepackage{color}
\usepackage{comment}
\usepackage{float}
\usepackage{gensymb}
\usepackage{graphicx}
\usepackage[colorlinks]{hyperref}
\usepackage{hyperref}
\usepackage{mathtools}
\usepackage{multirow}
\usepackage{nicefrac}
\usepackage[caption=false,font=footnotesize,subrefformat=parens,labelformat=parens]{subfig}
\usepackage{textcomp}
\usepackage{tikz}
\usepackage{wrapfig}
\hypersetup{
    linkcolor={red!95!white},
    citecolor={green!95!white},
    urlcolor={magenta!95!white}
}
\graphicspath{{images/}}

\newcommand\copyrighttext{%
  \footnotesize \textcopyright 2021 IEEE. Personal use of this material is
  permitted. Permission from IEEE must be obtained for all other uses, in any
  current or future media, including reprinting/republishing this material for
  advertising or promotional purposes, creating new collective works, for resale
  or redistribution to servers or lists, or reuse of any copyrighted component of
  this work in other works. DOI: \href{https://doi.org/10.1109/ICUAS51884.2021.9476712}{10.1109/ICUAS51884.2021.9476712}}
\newcommand\copyrightnotice{%
\begin{tikzpicture}[remember picture,overlay]
\node[anchor=south,yshift=10pt] at (current page.south) {\fbox{\parbox{\dimexpr\textwidth-\fboxsep-\fboxrule\relax}{\copyrighttext}}};
\end{tikzpicture}%
}

\newcommand{\at}[2][]{#1|_{#2}}

\newcommand{\simplotwidth}{0.45\textwidth}
\definecolor{BLOOD_RED}{RGB}{110, 14, 10}
\definecolor{YELLOW_ORANGE}{RGB}{255,167,55}
\definecolor{GREEN_PANTONE}{RGB}{77,170,87}

\begin{document}

\title{Vision-Based Guidance for Tracking Dynamic Objects}

\author{Pritam Karmokar$^{1*}$, Kashish Dhal$^{2*}$, 
        William J. Beksi$^{1}$, and Animesh Chakravarthy$^{2}$ 
\thanks{*Indicates equal contribution}%
\thanks{$^{1}$ P. Karmokar and W.J. Beksi are with the Department of Computer 
        Science and Engineering, University of Texas at Arlington, Arlington, TX,
        USA. 
        Emails: 
        pritam.karmokar@mavs.uta.edu,
        william.beksi@uta.edu
        }
\thanks{$^{2}$ K. Dhal and A. Chakravarthy are with the Department of Mechanical 
        and Aerospace Engineering, University of Texas at Arlington, Arlington, TX, 
        USA. 
        Emails: 
        kashish.dhal@mavs.uta.edu,
        animesh.chakravarthy@uta.edu
        }
}

\maketitle
\copyrightnotice
\pagestyle{plain}

\begin{abstract}
In this paper, we present a novel vision-based framework for tracking dynamic
objects using guidance laws based on a rendezvous cone approach. These guidance
laws enable an unmanned aircraft system equipped with a monocular camera to
continuously follow a moving object within the sensor's field of view. We
identify and classify feature point estimators for managing the occurrence of
occlusions during the tracking process in an exclusive manner. Furthermore, we
develop an open-source simulation environment and perform a series of
simulations to show the efficacy of our methods.  
\end{abstract}


\section{Introduction}
\label{sec:introduction}
In recent years there has been an increase in the number of applications using
unmanned aircraft systems (UASs). At the same time, researchers have
progressively inclined towards using vision as a primary source of perception
\cite{kanellakis2017survey}. This is mainly due to cameras becoming cheaper in
cost, smaller in size, lighter in weight, and higher in image resolution.
Likewise, as computing resources evolve to be more economical and powerful,
there has been growing interest in research and development for UASs. On account
of their agility, mobility, and form factor, various diverse problems have found
UASs with on-board vision-based sensors to be an ideal solution
\cite{al2018survey}. Not only can UASs reach places that are intractable for
humans to access, but they are also excellent platforms for monotonous and
dangerous jobs. This includes traffic monitoring
\cite{kanistras2013survey,zhou2014efficient}, search and rescue
\cite{yeong2015review,van2017first}, reconnaissance for military operations
\cite{samad2007network,manyam2017multi}, and much more. 

\begin{figure}
\centering
\includegraphics[width=0.45\textwidth]{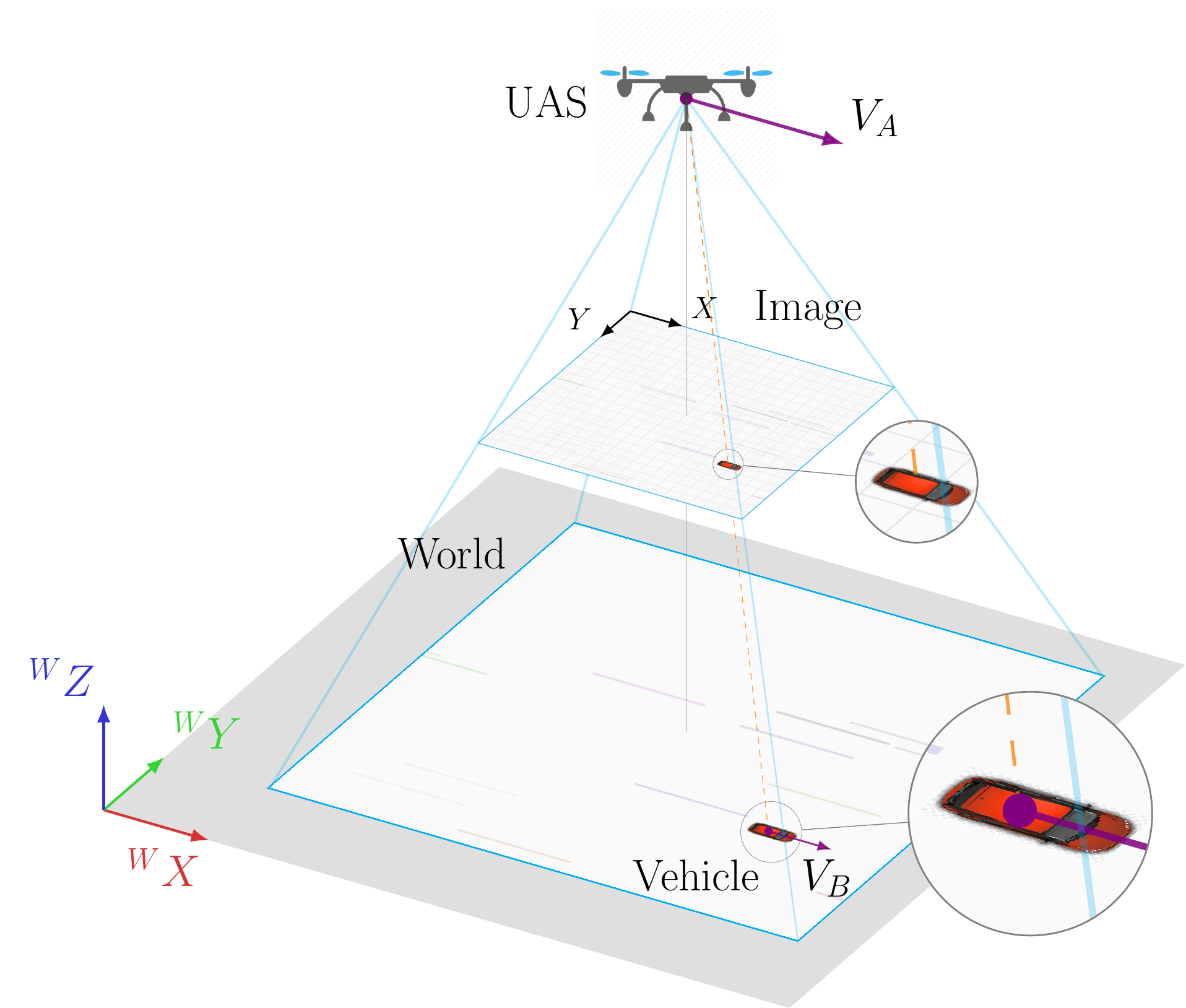}
\caption{Our proposed framework allows an unmanned aircraft system to visually
track a ground vehicle under the existence of partial and total occlusions using
unique guidance laws based on a rendezvous cone approach.}
\label{fig:uas_vehicle}
\vspace{-2mm}
\end{figure}

In this work, we construct a system that permits a UAS to pursue a dynamic
ground vehicle using only visual information. To do this we employ the concept
of a rendezvous cone. Concretely, we analytically show how a rendezvous cone can
be used to develop guidance laws for a UAS to track a moving vehicle. These
guidance laws are supported by a comprehensive set of computer vision algorithms
that perform feature detection, track and make adjustments to the centroid of
the vehicle, and filter for robustness and recovery through partial and full
occlusions. In summary, our key contributions are the following.
\begin{itemize}
  \item We introduce a set of systematic guidance laws which allow a UAS to
  visually track a moving object.
  \item We provide an identification and taxonomy of occlusion handling for
  feature point estimators under various occlusion cases.
  \item We implement an extensive open-source simulator
  \cite{a_vision-based_guidance_for_tracking_dynamic_objects} that supplies
  visual measurements while performing object tracking. 
\end{itemize}
Moreover, our proposed system can be applied to applications such as tracking,
monitoring, and surveillance via UASs.

The remainder of this paper is organized as follows. We discuss related work in
Section~\ref{sec:related_work}. Section~\ref{sec:problem_statement} provides a
concise statement of the problem to be solved. Our vision-based guidance scheme
is presented in Section~\ref{sec:vision_based_guidance}. In
Section~\ref{sec:simulations}, we report our simulation results. We conclude in
Section~\ref{sec:conclusion} and mention future work.

\section{Related Work}
\label{sec:related_work}
\subsection{Cones in Relative Velocity Space}
There has been related work on the development of guidance laws using cones
constructed in the relative velocity space.  The concept of collision cones was
proposed in \cite{chakravarthy1998obstacle} to represent a collection of
velocity vectors of an object which leads to a collision with another moving
object. Guidance laws to avoid collision were then designed to steer the current
velocity vector of the object outside the collision cone. This idea was later
employed by many researchers for various applications ranging from aircraft
conflict detection and resolution \cite{goss2004aircraft}, vision-based obstacle
avoidance \cite{watanabe2006minimum,watanabe2007vision}, automobile collision
avoidance \cite{ferrara2009second}, robotic collision avoidance
\cite{gopalakrishnan2014time}, and to study collision avoidance behavior in
biological organisms \cite{brace2016using}. 

In subsequent work, collision cones have been extended to higher-dimensional
spaces and general obstacle shapes \cite{chakravarthy2017collision}. They've
also been used to design safe passage trajectories through narrow orifices for
aerial as well as underwater vehicles \cite{zuo2020model}. In contrast to these
prior research problems, the guidance laws presented in our work specify
accelerations which when applied by a UAS enables it to steer its velocity
vector {\em towards} the moving object and subsequently match its velocity to
that of the object. The term rendezvous cone is more representative for this
class of applications and we shall adopt this term in our presentation. In
addition, our rendezvous cone approach employs vision-based information in order
to perform its computations.       

\subsection{Vision-Based Tracking}
The first comprehensive survey on visual tracking and categorization of
state-of-the-art algorithms was presented in \cite{yilmaz2004object}. Further
research was provided by \cite{yilmaz2006object} and \cite{porikli2012object}
where the main trends and taxonomy in object detection and tracking are
introduced. Various vision-based tracking techniques exist ranging from
low-level HSV threshold color-based detection
\cite{carelli2005vision,lee2016vision} through high-level tracking-by-detection
learning-based methods
\cite{henriques2014high,bergmann2019tracking,zhou2020tracking}. To overcome
known challenges in visual tracking, many ideas such as template matching
\cite{pan2007robust}, feature matching \cite{shen2013vision}, and optical flow
\cite{madasu2010estimation,li2019siamrpn++,wang2019fast} have been adopted by
researchers.

Our work focuses on visual object tracking
\cite{dobrokhodov2006vision,jeon2013mode,kim2014visual,liu2019vehicle} utilizing
a point estimator for target localization. In addition, we perform occlusion
handling \cite{lee2014occlusion} and identify occlusion states
\cite{galton1994lines,kohler2002occlusion} along with case transitions for
tracking point estimators. Tools and methods such as part-based templates
\cite{zhang2014partial}, RGB color distributions \cite{bouachir2014structure},
segmentation \cite{han2009tracking}, deformable models \cite{pang2004novel},
background/foreground \cite{guha2011formulation} detection, etc., have been
employed to carry out occlusion management. In comparison, our approach makes
use of classical computer vision techniques \cite{zhou2019does}. Specifically,
we implement a resilient feature detection strategy for tracking and adjusting
the centroid of an object in the presence of extreme occlusions.

\section{Problem Statement}
\label{sec:problem_statement}
Consider a scenario in which a UAS, equipped with a downward-facing camera, is
flying at a known altitude. In addition, the UAS only perceives visual
information as it tracks a ground-based vehicle. It is assumed the vehicle has
been detected and is initially within the UAS's field of view. Moreover, the
motion of the vehicle is along a plane orthogonal to the principal axis of the
camera and thus the image projection is orthographic. The problem is to develop
vision-based guidance laws that facilitate the UAS in consistently tracking the
vehicle even as it performs evasive maneuvers and encounters occlusions.  

\section{Vision-Based Guidance}
\label{sec:vision_based_guidance}

\subsection{Engagement Geometry and Guidance Laws}
\label{subsec:engagement_geometry_and_guidance_laws}
\begin{figure}
\centering
\includegraphics[width=0.4\textwidth]{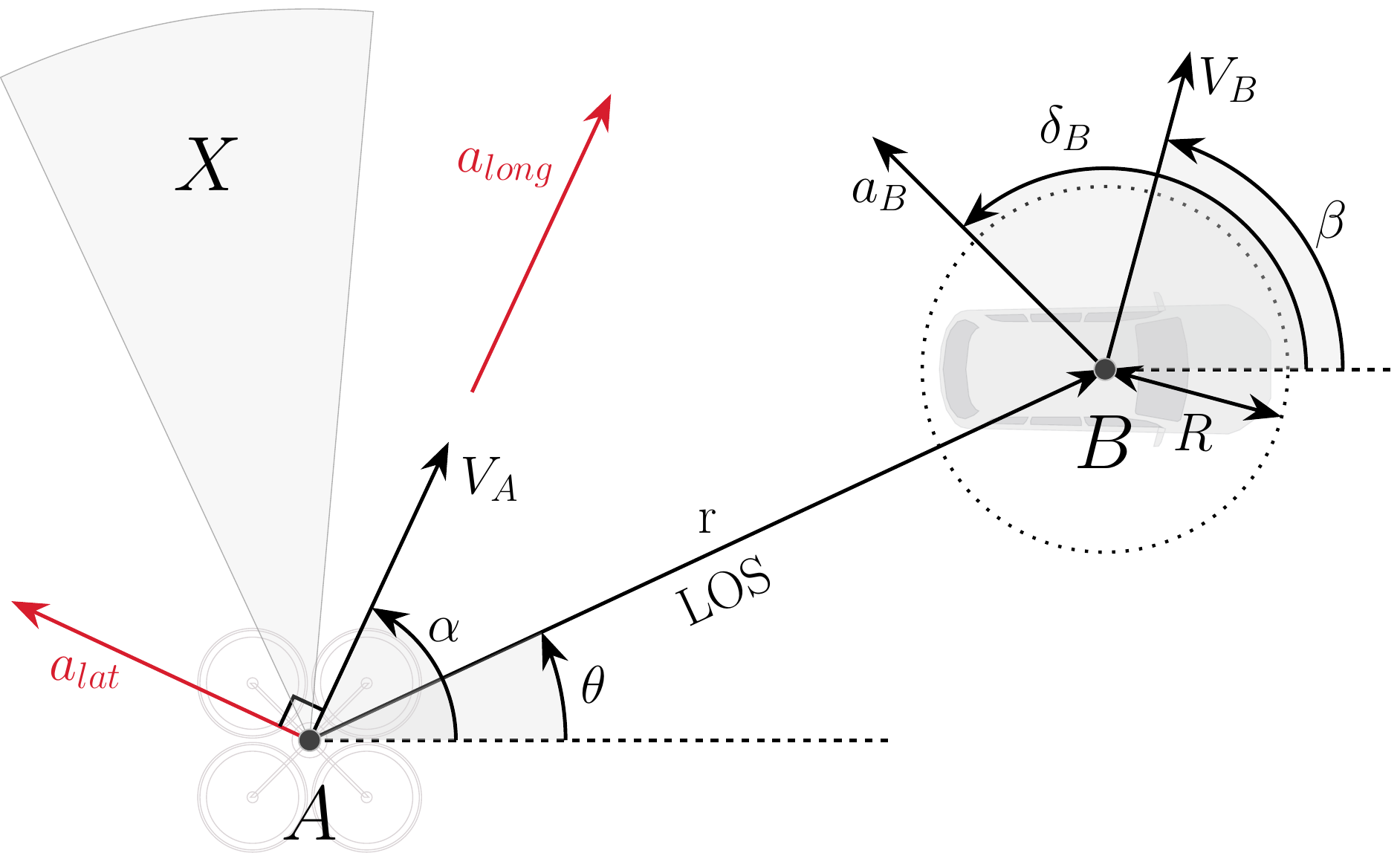}
\caption{An engagement between a UAS ($A$) and a ground vehicle ($B$).}
\label{fig:engagement_geometry}
\end{figure}

We model the engagement geometry between a UAS ($A$) and vehicle ($B$) as an
orthogonal projection onto a horizontal plane,
Fig.~\ref{fig:engagement_geometry}. The UAS and vehicle are moving with speeds
$V_A$ and $V_B$, and heading angles $\alpha$ and $\beta$, respectively. The
distance between $A$ and $B$ is represented by $r$ while $\theta$ denotes the
angle made by the line of sight (LOS) $AB$. The UAS has two control inputs:
lateral acceleration $a_{lat}$ and longitudinal acceleration $a_{long}$. Thus,
$A$ can rotate its velocity vector as well as change its speed. The vehicle can
apply an acceleration of magnitude $a_B$ which acts at an angle $\delta_B$, and
$R$ denotes the radius of $B$. $X$ portrays the rendezvous cone from $A$ to $B$.
If $A$ can steer its velocity vector into $X$, then $A$ is on a trajectory that
will cause it to rendezvous with $B$. The kinematics governing the engagement
geometry are characterized by the equations governing $AB$, 
\begin{eqnarray}
  \begin{bmatrix} \label{eq:state_kinematics}
    \Dot{{r}}\\
    \Dot{{\theta}}\\
    \Dot{{V_\theta}}\\
    \Dot{{V_r}}\\
    \Dot{\alpha}\\
    \Dot{V_A}\\
    \Dot{\beta}\\
    \Dot{{V_B}}
  \end{bmatrix}
  &=&
  \begin{bmatrix}
    {V_r}\\
    {{V_\theta/r}}\\
    -{V_\theta}{V_r}/{r}\\
    V_\theta ^2/{r}\\
    0\\
    0\\
    0\\
    0
  \end{bmatrix}
  + 
  \begin{bmatrix}
    0\\
    0\\
    -\cos(\alpha - {\theta})\\
    \sin(\alpha - {\theta})\\
    1/V_A\\
    0\\
    0\\
    0
  \end{bmatrix}a_{lat}\\
   &+&
  \begin{bmatrix} 
    0\\
    0\\
    -\sin(\alpha- {\theta})\\
    -\cos(\alpha - {\theta})\\
    0 \\
    1 \\
    0 \\
    0
  \end{bmatrix}a_{long}
  + \begin{bmatrix} 
    0\\
    0\\
    \sin(\delta_B - {\theta})\\
    \cos(\delta_B - {\theta})\\
    0\\
    0\\
    \sin(\delta_B - {\beta})/V_B\\
    \cos(\delta_B - {\beta})
  \end{bmatrix}a_{B},\nonumber
\end{eqnarray}
where $V_r$ and $V_\theta$ are the components of the relative velocity vector.   

We define the rendezvous cone \cite{chakravarthy1998obstacle} as 
\begin{equation}
\label{y_eqn1}
  y_1 = r^2 V_\theta^2 - R^2 (V_\theta^2 + V_r^2),
\end{equation}
i.e., it is the cone of relative velocity vectors that will cause $A$ to
rendezvous with $B$. This is established by two conditions: (i) $y_1<0,V_r<0$
and (ii) $y_1=0,V_r<0$. Condition (i) corresponds to the case of the relative
velocity vector being inside the rendezvous cone while condition (ii)
corresponds to the scenario of the relative velocity vector being aligned with
the boundary of the rendezvous cone. When (ii) occurs $A$ will graze $B$ at the
instant of closest approach. We define the velocity matching error as
\begin{equation}
\label{y_eqn2}
  {y}_2 = V_r^2 + V_\theta^2.
\end{equation}

Dynamic inversion is employed to drive the output functions to the desired
values. By differentiating \eqref{y_eqn1} and \eqref{y_eqn2} we obtain the
dynamic evolution of $y_1$ and $y_2$ as
\begin{equation}
  \label{eq:yder}
  \begin{bmatrix} 
    \dot{y}_1\\
    \dot{y}_2
  \end{bmatrix} =  
  \begin{bmatrix}  
    \frac{\partial y_1}{\partial r} & \frac{\partial y_1}{\partial \theta} & \frac{\partial y_1}{\partial V_{\theta}} & \frac{\partial y_1}{\partial V_{r}}\\ 
    \frac{\partial y_2}{\partial r} & \frac{\partial y_2}{\partial \theta} & \frac{\partial y_2}{\partial V_{\theta}} & \frac{\partial y_2}{\partial V_{r}} 
   \end{bmatrix}
  \times 
  \left[ \begin{array}{c}{\dot{r}} \\ {\dot{\theta}} \\ {\dot{V}_{\theta}} \\ {\dot{V}_{r}} \\ \end{array}\right].
\end{equation}
The partial derivatives of \eqref{eq:yder} are expressed as 
\begin{equation}
\begin{bmatrix}  
  \label{eq:ypartial}
  2rV_{\theta}^2 & 0 & 2 V_{\theta}(r^2-R^2) & -2r^2V_r\\ 
               0 & 0 & 2 V_{\theta}          & 2V_r
\end{bmatrix}.
\end{equation}

To calculate the required control input, we define two error quantities with
respect to $y_1$ and $y_2$. The error in $y_1$ is specified as $e_1(t) =
y_{1d}(t) - y_1(t)$, where $y_{1d}(t)<0$ is a reference input while the error in
$y_2$ is described as $e_2(t)=0-y_2(t)$. Taking $y_{1d}(t)$ as a constant
$\forall t$, we seek to determine $a_{lat}$ and $a_{long}$ which will ensure the
error dynamics follow the equations $\dot{e}_1 = -k_1e_1$ and $\dot{e}_2=-k_2
e_2$ where $k_1,k_2 >0$ are constants. This in turn allows the quantities $y_1$
and $y_2$ to follow the dynamics, i.e.,
\begin{eqnarray}
\begin{bmatrix} 
  \dot{y}_1\\
  \dot{y}_2
\end{bmatrix}
&=&
\begin{bmatrix} 
  k_1(y_{1d}-y_1)\\
  -k_2y_2
\end{bmatrix}.
\label{eq:y1_y2_dot}
\end{eqnarray}
After substituting \eqref{eq:state_kinematics}, \eqref{eq:ypartial}, and
\eqref{eq:y1_y2_dot} into \eqref{eq:yder} we obtain 
\begin{eqnarray}
\begin{bmatrix} 
  a_{11} & a_{12}\\
  a_{21} & a_{22}
\end{bmatrix}
\begin{bmatrix} 
  a_{lat}\\
  a_{long}
\end{bmatrix}
  =\begin{bmatrix} 
  -k_1(y_{1d} - y_1)/2\\
  k_2y_2/2
\end{bmatrix} - 
\begin{bmatrix} 
  b_1\\
  b_2
\end{bmatrix}
a_{B},
\label{eq:alat_along_eqn}
\end{eqnarray}
where
\begin{eqnarray}
  a_{11} &=& V_{\theta} ( r^2-R^2 ) \cos(\alpha-\theta) + R^2V_r\sin(\alpha-\theta), \nonumber\\
  a_{12} &=& V_{\theta} ( r^2-R^2 ) \sin(\alpha-\theta) + R^2V_r\cos(\alpha-\theta), \nonumber\\
  a_{21} &=& V_{\theta}  \cos(\alpha-\theta) + V_r\sin(\alpha-\theta), \nonumber\\
  a_{22} &=& V_{\theta} \sin(\alpha-\theta) + V_r\cos(\alpha-\theta), \nonumber\\
  b_1 &=& V_r R^2 \cos(\delta_B-\theta) - V_{\theta}(r^2-R^2) \sin(\delta_B-\theta), \nonumber\\
  b_2 &=&  - V_r \cos(\delta_B-\theta)  -V_{\theta} \sin(\delta_B-\theta). \nonumber 
\end{eqnarray}
%
%
By solving \eqref{eq:alat_along_eqn} we obtain 
\begin{align}
  a_{lat} &= \Big[ k_1 (y_1-y_{1d}) \Big(V_r  \cos(\alpha - \theta)  +  V_\theta  \sin(\alpha - \theta) \Big) \nonumber \\
    &+ k_2 y_2 \Big(V_r R^2    \cos(\alpha - \theta) -  V_\theta (r^2-R^2)  \sin(\alpha - \theta) \Big) \nonumber\\
    &- 2V_r V_\theta r^2 \sin(\alpha-\delta_B) a_B \Big] /(2 V_r V_\theta r^2) \label{eq:a_lat_final}\\
  a_{long} &= \Big[  k_1 (y_1-y_{1d}) \Big( V_r  \sin(\alpha - \theta) -  V_\theta \cos(\alpha - \theta)\Big) \nonumber\\
    &+ k_2 y_2 \Big(V_r  R^2 \sin(\alpha - \theta) +  V_\theta (r^2-R^2)   \cos(\alpha - \theta) \Big) \nonumber\\
    &+ 2V_r V_\theta r^2 \cos(\alpha-\delta_B) a_B \Big] /(2 V_r V_\theta r^2). \label{eq:a_long_final}
\end{align}

\eqref{eq:a_lat_final} and \eqref{eq:a_long_final} serve as the acceleration
commands to the UAS and are supplied to our simulator. We note the acceleration
commands $a_{lat}(t)$ and $a_{long}(t)$ depend on continuous feedback from the
quantities $y_1$ and $y_2$, states $r,\theta,V_r,V_\theta$, and the acceleration
vector of the vehicle. The quantities $r$ and $\theta$ are obtained by direct
measurements from the vision system while the remaining quantities are estimated
using an acceleration model (Section~\ref{subsec:vehicle_acceleration_model}).

\subsection{Vehicle Acceleration Model} 
\label{subsec:vehicle_acceleration_model}
The positional measurements of the vehicle in the image frame are obtained using
feature point tracking (Section~\ref{subsec:feature_tracker}). In addition,
since it's assumed the UAS knows its own states, we can transform the image
measurements to an inertial frame. These inertial measurements are then fed into
a linear acceleration model \cite{singer1970estimating,mahapatra2000mixed}. The
model filters the position data and provides estimates of the velocity and
acceleration of the vehicle. Given that the position of the vehicle in the
inertial frame is represented as $(x_B,y_B)$, we obtain the discrete form of the
motion model as
\begin{eqnarray}
  \label{eq:motion_model}
  \bm{\chi}_B(k) &=& \bm{F}_B \bm{\chi}_B(k-1) + \bm{w}_B(k), \nonumber\\
  \bm{z}_B(k)  &=& \bm{H}_B\bm{\chi}_B(k) + \bm{v}_B(k). 
\end{eqnarray}
where the state vector of the vehicle is represented by $\bm{\chi}_B =
[x_B,\dot{x}_B,\Ddot{x}_B,y_B,\dot{y}_B, \Ddot{y}_B]^\top$ and $\bm{z}_B$ is the
measurement vector. The state transition matrix corresponding to the
acceleration model is defined as
\begin{equation}
  \bm{F}_B =
    \begin{bmatrix}  
      \bm{T} & \bm{0}\\
      \bm{0} & \bm{T}
    \end{bmatrix}, \nonumber
\end{equation}
where $\bm{0}$ is a $3 \times 3$ matrix of zeros and
\begin{equation}
  \bm{T} = 
    \begin{bmatrix}  
      1 & \Delta t & [e^{-\alpha_B \Delta t}+ \alpha_B \Delta t -1]/\alpha_B^2\\
      0 & 1 & [1-e^{-\alpha_B \Delta t}]/\alpha_B\\
      0 & 0 & e^{-\alpha_B \Delta t}
  \end{bmatrix}. \nonumber
\end{equation}
The process noise is defined as $\bm {w}_B \sim {\mathcal{N}}{\bigl(}{0},
\bm{Q}_B{\bigr )}$ and the noise covariance matrix is given by
\begin{equation}
  \bm{Q}_B = 2 \alpha_B \sigma_B^2
    \begin{bmatrix}  
      \bm{Q} & \bm{0}\\
      \bm{0} & \bm{Q}
    \end{bmatrix}, \nonumber
\end{equation}
where $\alpha_B$, $\sigma_B$, and $\bm{Q}$ are described in
\cite{singer1970estimating}. The size of $\bm{Q}$ and $\bm{0}$ is $3 \times 3$.
The measurement matrix takes the form
\begin{equation}
  \bm{H}_B =
    \begin{bmatrix}  
      1 & 0 & 0 & 0 & 0 & 0\\
      0 & 0 & 0 & 1 & 0 & 0
    \end{bmatrix}. \nonumber
\end{equation}
$\bm{v}_B\sim {\mathcal{N}}{\bigl(}{0}, \bm{R}_B{\bigr)} $ is the measurement
noise and $\bm{R}_B$ is the measurement noise covariance matrix. The velocity
estimates $(\dot{x}_B,\dot{y}_B)$ of the vehicle are transformed back into the
UAS-centered relative frame to obtain the relative velocity components
$\hat{V}_r$ and $\hat{V}_\theta$. Similarly, the vehicle acceleration estimates
$(\ddot{x}_B,\ddot{y}_B)$ are translated into $\hat{a}_B$ and $\hat{\delta}_B$.

\subsection{System Architecture}
\begin{figure}
\centering
\includegraphics[width=0.45\textwidth]{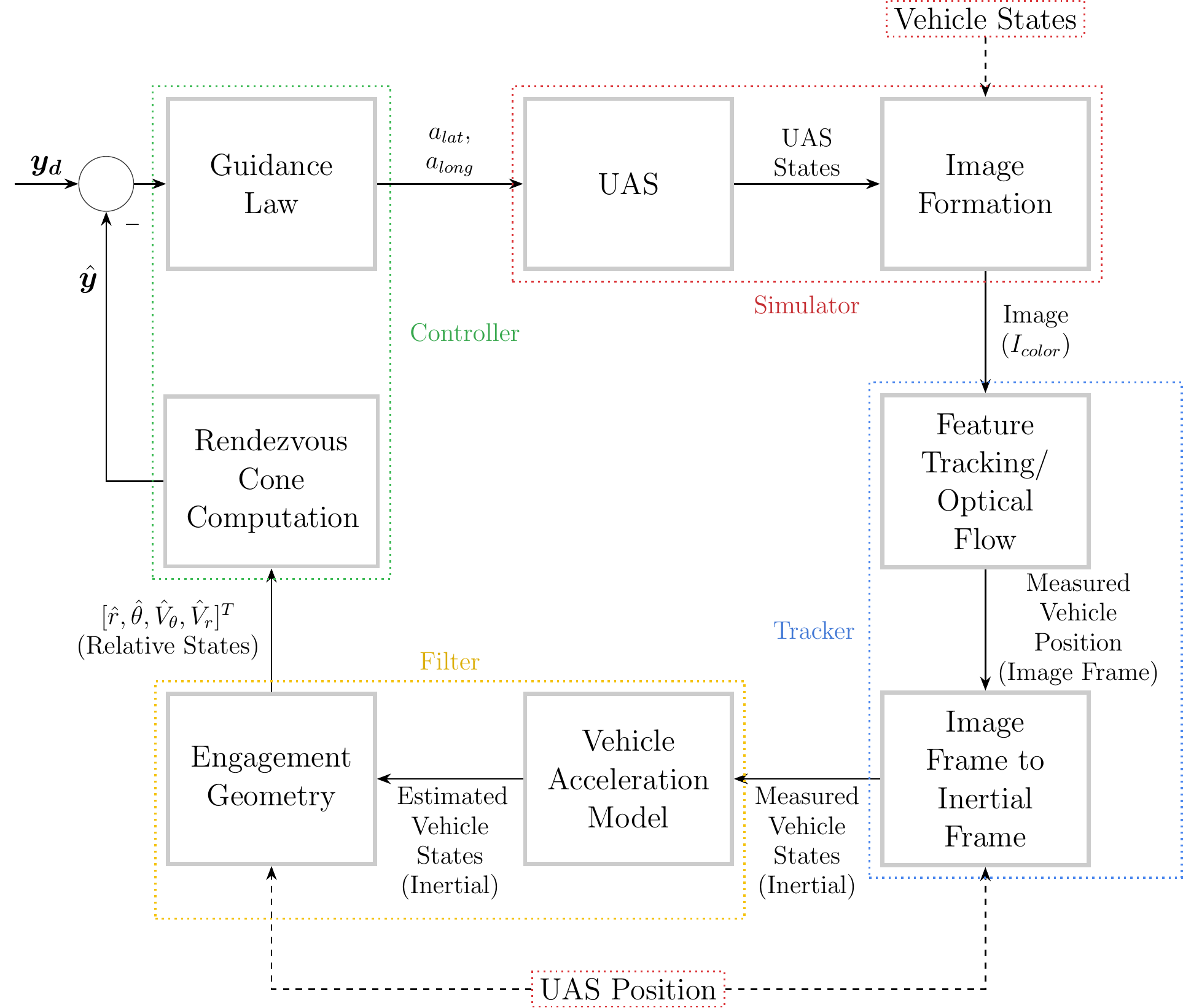}
\caption{The overall architecture of the vision-based guidance system.}
\label{fig:system_architecture}
\end{figure}
Our system architecture is shown in Fig ~\ref{fig:system_architecture}. First,
images captured by the UAS are used to estimate the relative and absolute
position of the vehicle. Next, the measured velocity and acceleration states are
processed to determine the quantities
$[\hat{r},\hat{\theta},\hat{V_\theta},\hat{V_r},\hat{a}_B,\hat{\delta}_B]^\top$.
The filtered estimates of the noisy $r_m$ and $\theta_m$ measurements are
represented by $\hat{r}$ and $\hat{\theta}$,  $\hat{V}_\theta$ and $\hat{V}_r$
express approximations of the relative velocity components, and $\hat{a}_B$ and
$\hat{\delta}_B$ represent estimates of the absolute acceleration magnitude and
direction of the vehicle. Finally, these values are used to compute an
assessment of the current rendezvous cone. An estimate of the UAS velocity
vector with respect to the rendezvous cone is denoted by $\hat{y}_1$, while
$\hat{y}_2$ measures the magnitude of the relative velocity vector. The guidance
algorithm generates suitable accelerations $a_{lat}$ and $a_{long}$ which drive
$\hat{y}_1$ and $\hat{y}_2$ to the desired reference values. These accelerations
move the UAS to a new position whereby an updated image of the scene is
generated by the simulator.  

\subsection{Simulator}
\label{subsec:simulator}
We have developed a Python Pygame \cite{pygame2021} simulator that constructs
orthographic projections from simulated kinematic states of a UAS and a vehicle.
The simulator generates RGB images at the appropriate frame rates or time
instances. Additionally, it can accept acceleration commands as input, update
the kinematic states of the simulated entities governed by the laws of rigid
body motion, and generate accurate renders. The simulator maps 3D scene points
onto a 2D image plane of RGB triplets. More formally, we describe this as the
composition of two mappings $\mathcal{M}_1: \mathbb{R}^{3} \rightarrow
\mathbb{P}^{2}$ and $\mathcal{M}_2: \mathbb{P}^{2} \rightarrow \mathbb{R}^{3}$
which yields $\mathcal{M}_2 \circ \mathcal{M}_1 : \mathbb{R}^{3} \rightarrow
\mathbb{R}^{3}$. The mapping $\mathcal{M}_1$ transforms kinematic states
${}^{W}\bm{x}(t) = \left[{}^{W}x(t),{}^{W}y(t),{}^{W}z(t)\right]^\top$ onto an
orthographic projective space. It generates the homogeneous vector
$\bm{\tilde{x}}(t) = \left[\tilde{x}(t), \tilde{y}(t), \tilde{w}(t)\right]^\top
= \tilde{w}\left[x(t),y(t),1\right]^\top$ where $\bm{x}(t) = [x(t),y(t)]^\top =
\left[\frac{\tilde{x}(t)}{\tilde{w}(t)},
\frac{\tilde{y}(t)}{\tilde{w}(t)}\right]^\top$. Conversely, the mapping
$\mathcal{M}_2$ discretizes and maps $\bm{\tilde{x}}(t)$ to image intensities
$I_{color}(x,y,t) = \left[r(t),g(t),b(t)\right]^\top$. The colored image is
converted to grayscale, $I(x,y,t)$, which is then consumed by the feature
tracker. It's worth mentioning that the simulator renders are non-anti-aliased
which introduces a discretization noise in the measurements since subpixel
motions are not simulated.

\subsection{Feature Tracker}
\label{subsec:feature_tracker}
\subsubsection{Optical Flow}
\label{subsubsec:optical_flow}
We obtain sparse optical flow at good and desirable feature points
\cite{shi1994good}. Feature points are good if the eigenvalues of the gradient
matrix at that point are large and they are desirable if they correspond to the
object to be tracked through a sequence of images. A Kanade-Lucas-Tomasi (KLT)
tracker \cite{lucas1981iterative,tomasi1991detection} is used for the task of
tracking feature point sets, $\mathcal{F}^k =
\left\{\bm{f}_i^k\right\}_{i=1}^{n}$, where $n$ is the number of points and
$\bm{f}_i^k = (x_i, y_i)^k$ is the $i^{th}$ point in the $k^{th}$ frame.

Optical flow displacements $\bm{u} \at {\bm{f}_i \in \mathcal{F}}$, where
$\bm{u} = \left[u_{x}, u_{y}, 1\right]^\top$, are obtained under the assumption
of a linearized brightness constancy \cite{horn1981determining},
\begin{equation} 
  \label{eq:optical_flow}
  I(x,y,t) = I(x+u_{x}, y+u_{y}, t+1).
\end{equation}
The displacements are computed over a small neighborhood
$\mathbf{\mathcal{N}}_{\bm{f}_i}$ of each $\bm{f}_i$. This is done by solving a
system of linear equations, 
\begin{equation} 
  \label{eq:linearized_optical_flow}
  \left(\nabla I\left(\bm{x}\right)\right)^{\top}\cdot\bm{u} = \mathbf{0}, \quad \forall \bm{x} \in \mathcal{N}_{\bm{f}},
\end{equation}
where $\nabla = \left[\frac{\partial}{\partial x}, \frac{\partial}{\partial y},
\frac{\partial}{\partial t}\right]^\top$ is the gradient operator. Using the
computed displacement vectors, updated locations $\bm{f}_i^{k+1} = (x_i,
y_i)^{k+1}$ are obtained. Furthermore, a multiscale pyramidal
\cite{memin1998multigrid,brox2004high,bruhn2005towards} approach enables
tracking of an object through larger displacements across frames.

\subsubsection{Centroid Adjustment}
\label{subsubsec:centroid_adjustment}
\begin{figure}
\centering \hfill
\subfloat[]{
  \includegraphics[width=0.2\textwidth]{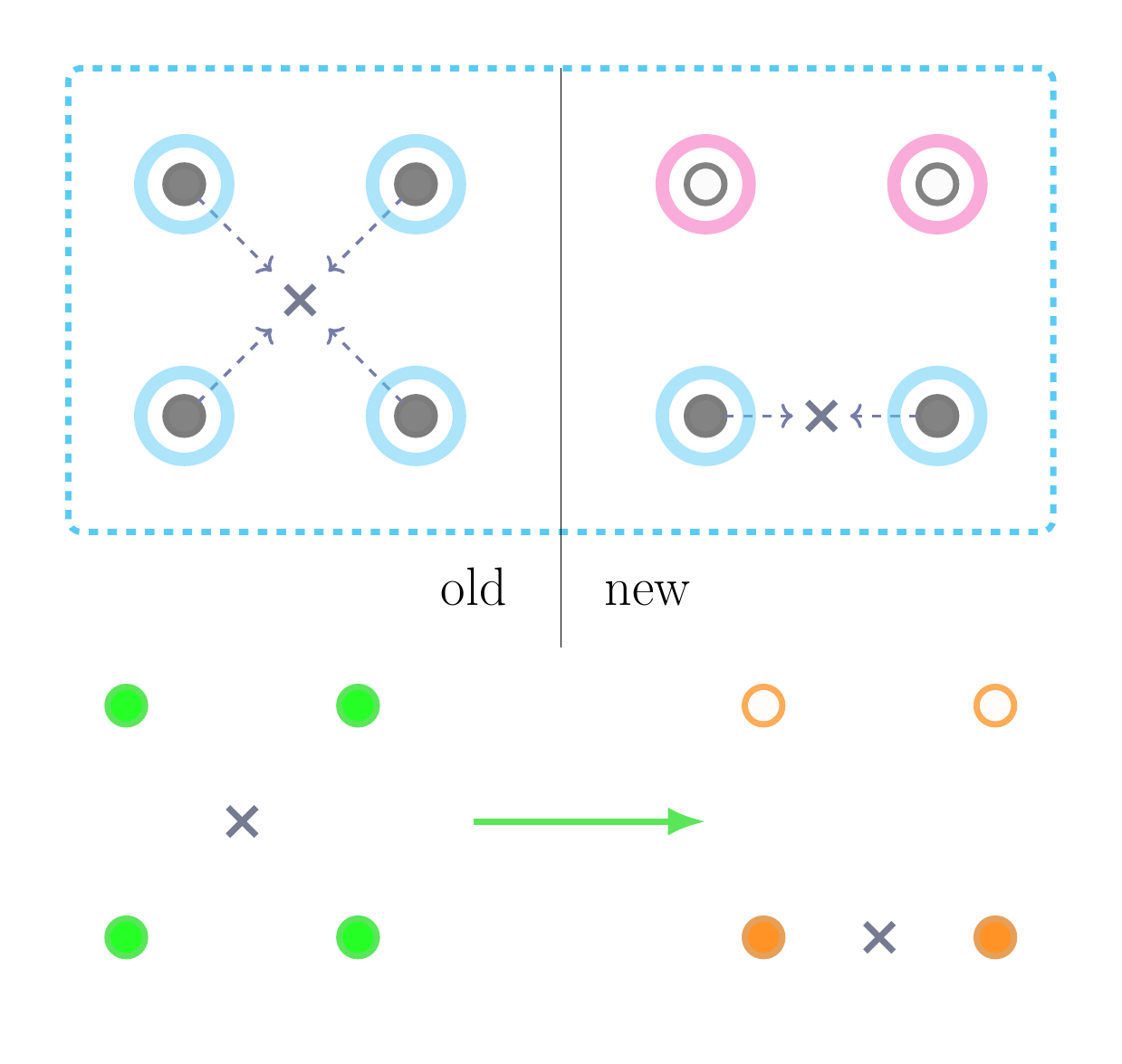}
  \label{fig:no_centroid_adjustment}
}\hfill
\subfloat[]{
  \includegraphics[width=0.2\textwidth]{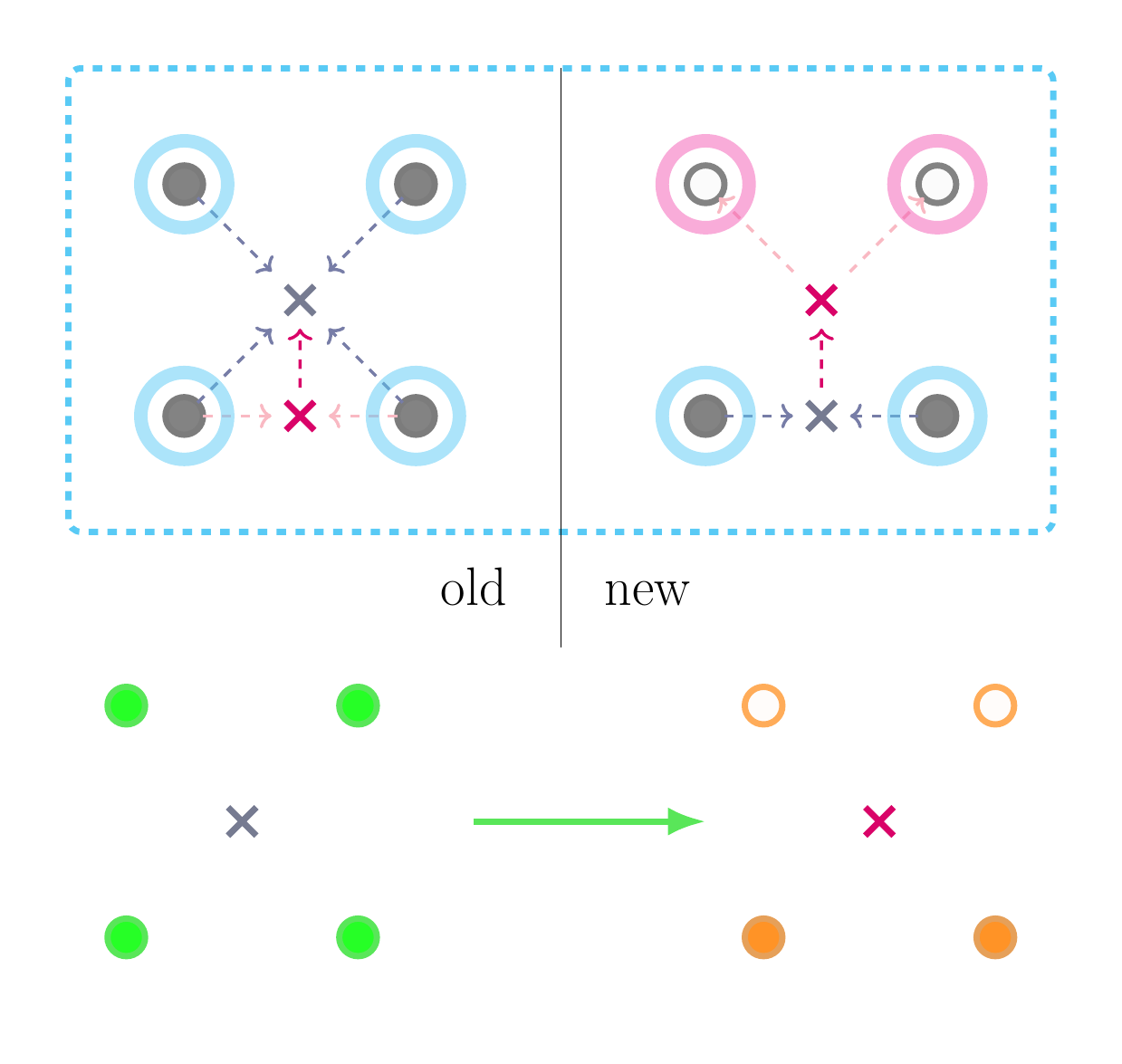}
  \label{fig:cent_adj}
}\hfill
\caption{Feature tracking for \textit{no occlusion} to \textit{partial
occlusion} state transitions: (a) without \textit{centroid adjustment}, (b) with
\textit{centroid adjustment}. \protect\tikz[baseline]
\protect\filldraw[draw=green!85!black!65,fill=green!85,line width=1.5pt]
(0ex,0.5ex) circle (0.75ex);-no occlusion, \protect\tikz[baseline]
\protect\filldraw[draw=orange!85!black!65,fill=orange!85,line width=1.5pt]
(0ex,0.5ex) circle (0.75ex);-partial occlusion, \protect\tikz[baseline]
\protect\draw[color=cyan!65,semitransparent,line width=2pt] (0ex,0.5ex) circle
(0.75ex);-good points (flow success), \protect\tikz[baseline]
\protect\draw[color=magenta!65,semitransparent,line width=2pt] (0ex,0.5ex)
circle (0.75ex);-bad points (flow failure), \protect\tikz[baseline]
\protect\filldraw[draw=darkgray!85!black!65,fill=darkgray!65,line width=1.5pt]
(0ex,0.5ex) circle (0.75ex);-visible points, \protect\tikz[baseline]
\protect\filldraw[draw=darkgray!65,fill=darkgray!2,line width=1.5pt] (0ex,0.5ex)
circle (0.75ex);-occluded points, \protect\tikz [baseline,x=1.4ex,y=1.4ex,line
width=1.5pt, gray] \protect\draw (-0.1,-0.1) -- (0.9,0.9) (-0.1,0.9) --
(0.9,-0.1);-not adjusted centroid, \protect\tikz [baseline,x=1.4ex,y=1.4ex,line
width=1.5pt, magenta] \protect\draw (-0.1,-0.1) -- (0.9,0.9) (-0.1,0.9) --
(0.9,-0.1);-adjusted centroid, \protect\tikz[baseline]
\protect\draw[color=cyan!65, line width=0.8pt, densely dotted] (0ex,-0.5ex)
rectangle (3ex,1.5ex);-optical flow across old and new frames.}
\label{fig:centroid_adjustment}
\end{figure}

Reliable optical flow computations require brightness constancy and temporal
regularity at feature points across frames. What's more, spatial consistency
such that pixels in the neighborhood of feature points have a motion similar to
the point, is needed as well. In practice, flow computations can be brittle when
factors such as nonlinear image distortions, noise, and occlusions occur.
Moreover, we may experience $m \le n$ missing points caused by high flow
optimization errors. This may result in unpredictable shifts in centroid
location despite the tracked object being relatively stationary. 

To mitigate these effects we implement centroid adjustment, i.e., the centroid
of $n$ feature points belonging to the object is localized for consistent
measurements. Formally, we define a centroid as 
\begin{equation}
  \label{eq:centroid}
  \bar{\mathcal{F}} = \frac{1}{n} \sum\limits_{i=1}^{n} \bm{f}_i.
\end{equation}
Adjustment of a centroid is then defined as 
\begin{align}
  \Delta{}^{}\bar{\mathcal{F}}^{}  &= {}^{\ast}\bar{\mathcal{F}}^{old} - {}^{good}\bar{\mathcal{F}}^{old},\\
  {}^{\ast}\bar{\mathcal{F}}^{new} &= {}^{good}\bar{\mathcal{F}}^{new} + \Delta{}^{}\bar{\mathcal{F}}^{},
\end{align}
where $\Delta{}^{}\bar{\mathcal{F}}^{}$ is the centroid adjustment and
${}^{\ast}\bar{\mathcal{F}}^{new}$ denotes the adjusted centroid.
Pre-superscripts are used to indicate good, bad, or all ($\ast$) points and
post-superscripts mean an old or new frame. This process is depicted in
Fig.~\ref{fig:centroid_adjustment}. 

Using centroid adjustment with the assumption of rigid shape constraints, if we
compute good flow for at least one keypoint ($m < n$), then we can obtain stable
centroid measurements. Furthermore, for resilience purposes, feature and
template matching are used as fallback techniques while handling occlusions.
Even though the object of interest is primarily represented using Shi-Tomasi
keypoints, we also have access to full and partial templates of the object. This
includes SIFT \cite{lowe2004distinctive} descriptors for redetection and
recovery purposes.

\subsection{Occlusion Handling}
\label{subsec:occlusion_handling}
In presence of occlusions, $\vert \mathcal{F} \vert = n$ cannot be assured.
Besides, $m > 0$ for at least $n$ time instances guarantees measurements will
vanish. In the context of tracking $n$ feature points
$\left\{\bm{f}_i\right\}_{i=1}^{n}$, we identify the following three occlusion
states ($\mathcal{O}$) with nine transitions among them: (i) no occlusion
($\mathcal{N}$), (ii) partial occlusion ($\mathcal{P}$), and (iii) total
occlusion ($\mathcal{T}$). The occlusion states can be detected and identified
by drawing inference from the cardinality $\left\vert {}^{good}\mathcal{F}
\right\vert$, i.e., 
\begin{equation}
\label{eq:occulsion_inference}
\begin{split}
  \vert {}^{good}\mathcal{F} \vert = n \quad & \implies \mathcal{N}\\
  0 < \vert {}^{good}\mathcal{F} \vert \le n \quad & \implies \mathcal{P}\\
  \vert {}^{good}\mathcal{F} \vert = 0 \quad & \implies \mathcal{T}.
\end{split}
\end{equation}

\begin{figure*}
\centering \hfill
\subfloat[]{
  \includegraphics[width=0.2\textwidth]{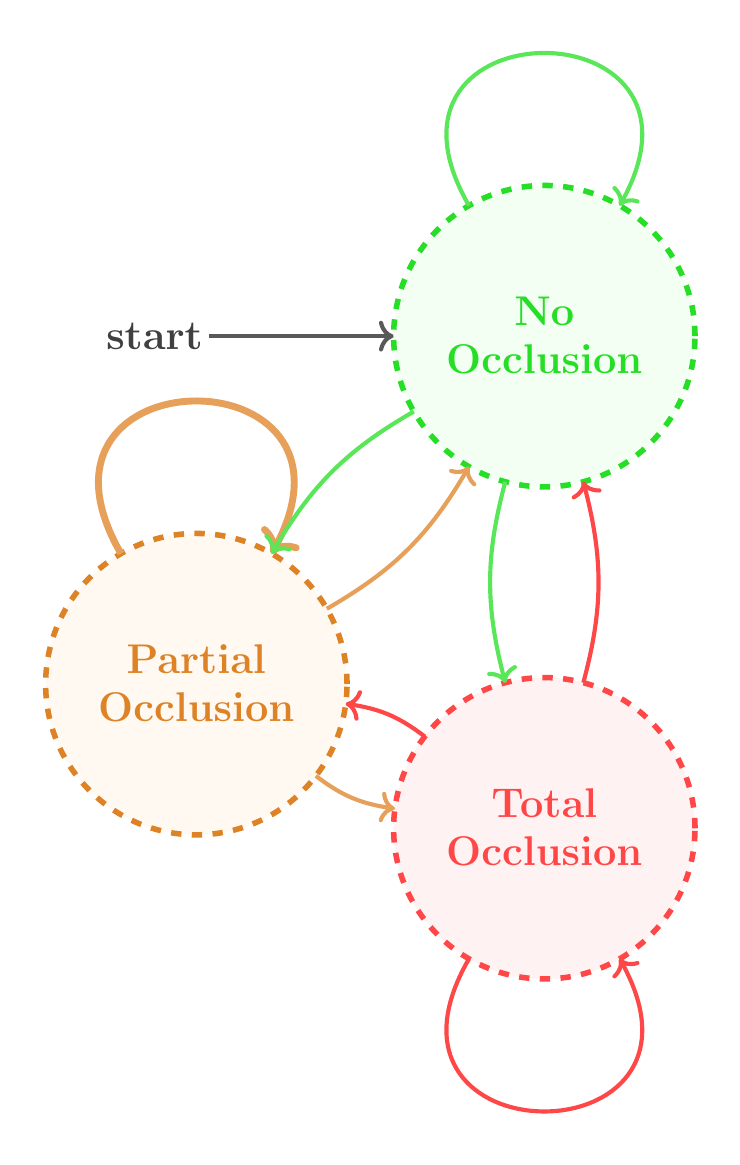}
  \label{fig:occulsion_state_transistions}
} \hfill
\subfloat[]{
  \includegraphics[height=0.125\textwidth,angle=270,origin=rb]{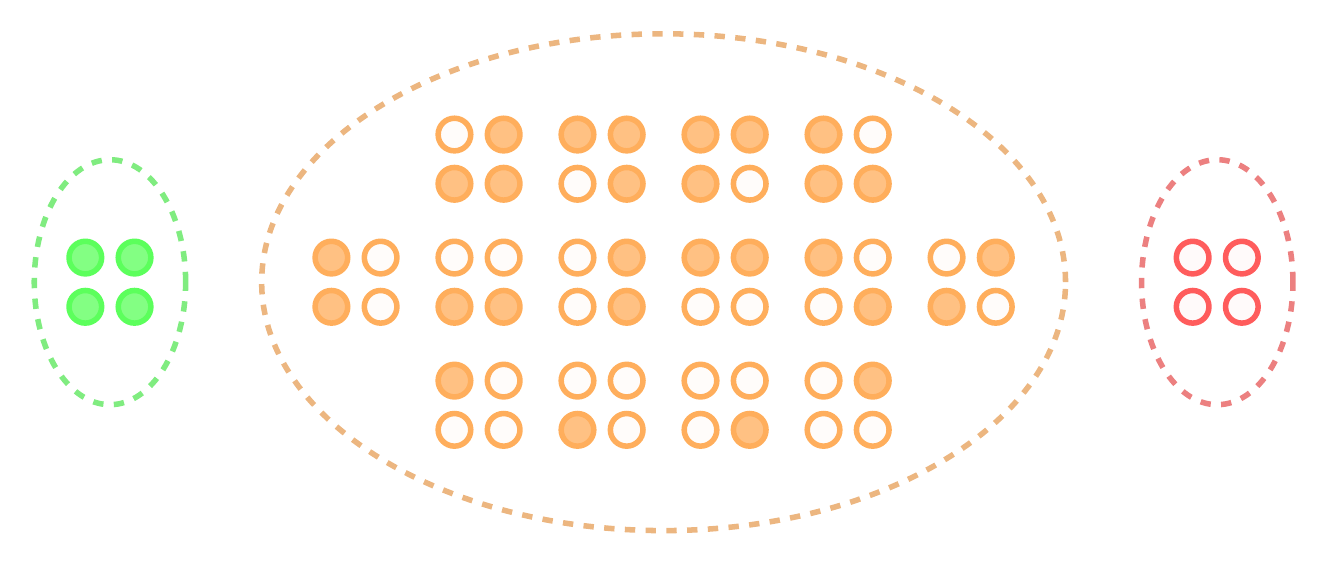}
  \label{fig:occulsion_state_sets}
} \hfill
\subfloat[]{
  \includegraphics[width=0.6\textwidth]{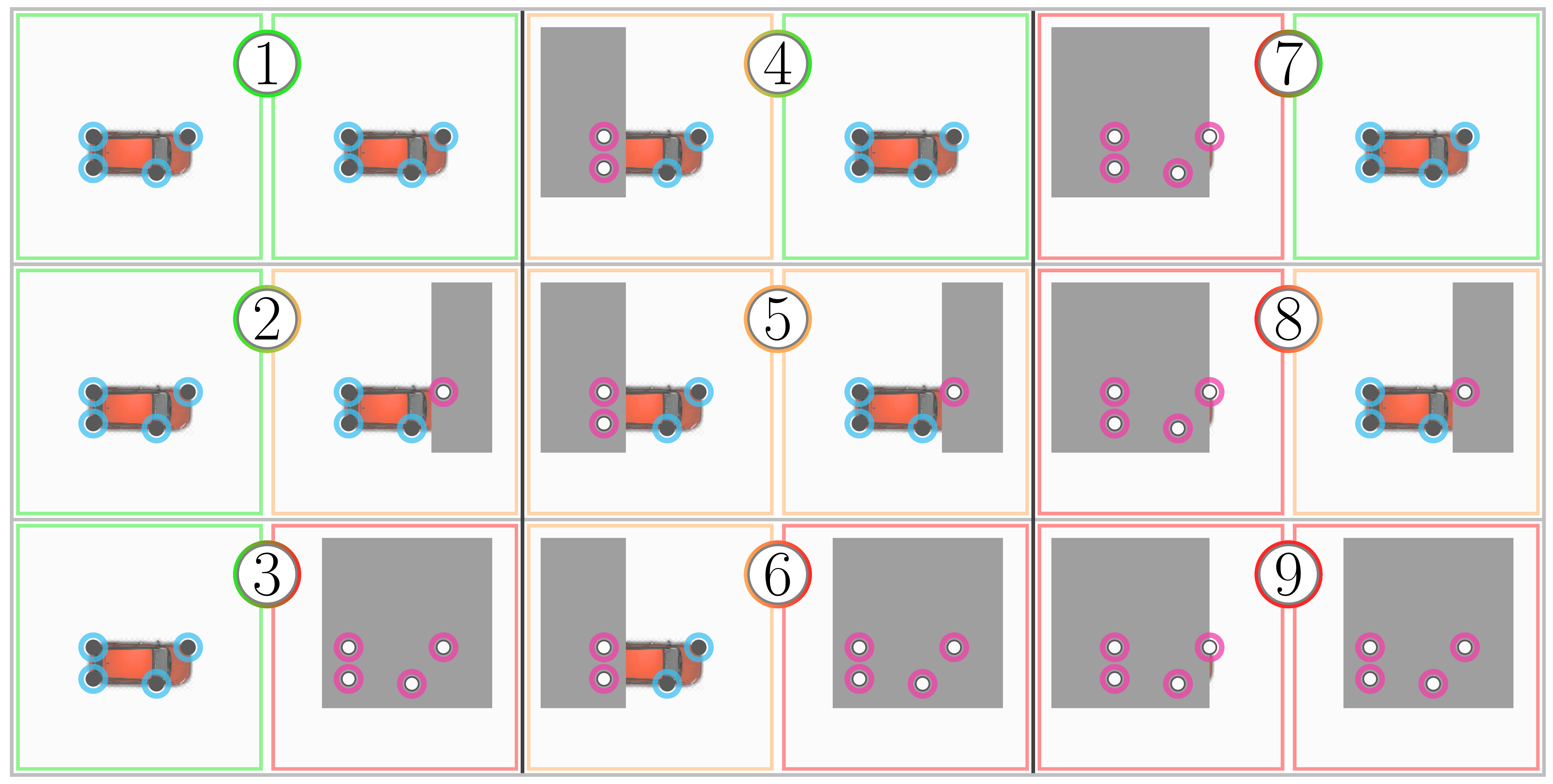}
  \label{fig:occulsion_cases}
} \hfill
\caption{(a) Occlusion state transitions. (b) Occlusion state sets:
\textcolor{green!85!black!85}{No Occlusion} \textit{(top)},
\textcolor{orange!85!black!85}{Partial Occlusion} \textit{(middle)},
\textcolor{red!85!white!85}{Total Occlusion} \textit{(bottom)}. (c) Enumerated
occlusion state transition case instances are depicted with a vehicle and
occlusion bars as follows: cases (1-3) enumerate transitions from \textit{no
occlusion} to \textit{no, partial, total occlusion}; cases (4-6) enumerate
transitions from \textit{partial occlusion} to \textit{no, partial, total
occlusion}; and cases (7-9) enumerate transitions from \textit{total occlusion}
to \textit{no, partial, total occlusion}.  \protect\tikz[baseline]
\protect\draw[color=cyan!65,semitransparent,line width=2pt] (0ex,0.5ex) circle
(0.75ex);-good points (flow success), \protect\tikz[baseline]
\protect\draw[color=magenta!65,semitransparent,line width=2pt] (0ex,0.5ex)
circle (0.75ex);-bad points (flow failure), \protect\tikz[baseline]
\protect\filldraw[draw=darkgray!85!black!65,fill=darkgray!65,line width=1.5pt]
(0ex,0.5ex) circle (0.75ex);-visible points, \protect\tikz[baseline]
\protect\filldraw[draw=darkgray!65,fill=darkgray!2,line width=1.5pt] (0ex,0.5ex)
circle (0.75ex);-occluded points.}
\end{figure*}

Although detecting states of occlusion is elementary, treating different
occlusion states in order to perform long-term point tracking via optical flow
is challenging. This is due to the fact that optical flow depends on the health
of the keypoints not only in the current frame, but also from the previous
frame. Consequently, to handle occlusions we enumerate nine transition instances
as shown in Fig.~\ref{fig:occulsion_cases}. Observe in
Fig.~\ref{fig:occulsion_state_sets} that the \textit{no occlusion} state set
$\mathcal{N}$ and the \textit{total occlusion} state set $\mathcal{T}$ are
singletons. Yet, the \textit{partial occlusion} state is a finite set
$\mathcal{P} = \left\{\mathcal{F} \mid \ 0 < \vert \mathcal{F} \vert \le n
\right\}$ with multiple elements. Accordingly, cases 1, 3, 7, 9 are the easiest
to handle, cases 2, 4, 6, 8 are moderate in complexity, and case 5 is the most
complex. A further breakdown of case 5 is discussed in
Section~\ref{subsubsec:from_partial_occlusion}.

\begin{figure}
\centering
\subfloat[]{
  \includegraphics[width=0.4\textwidth]{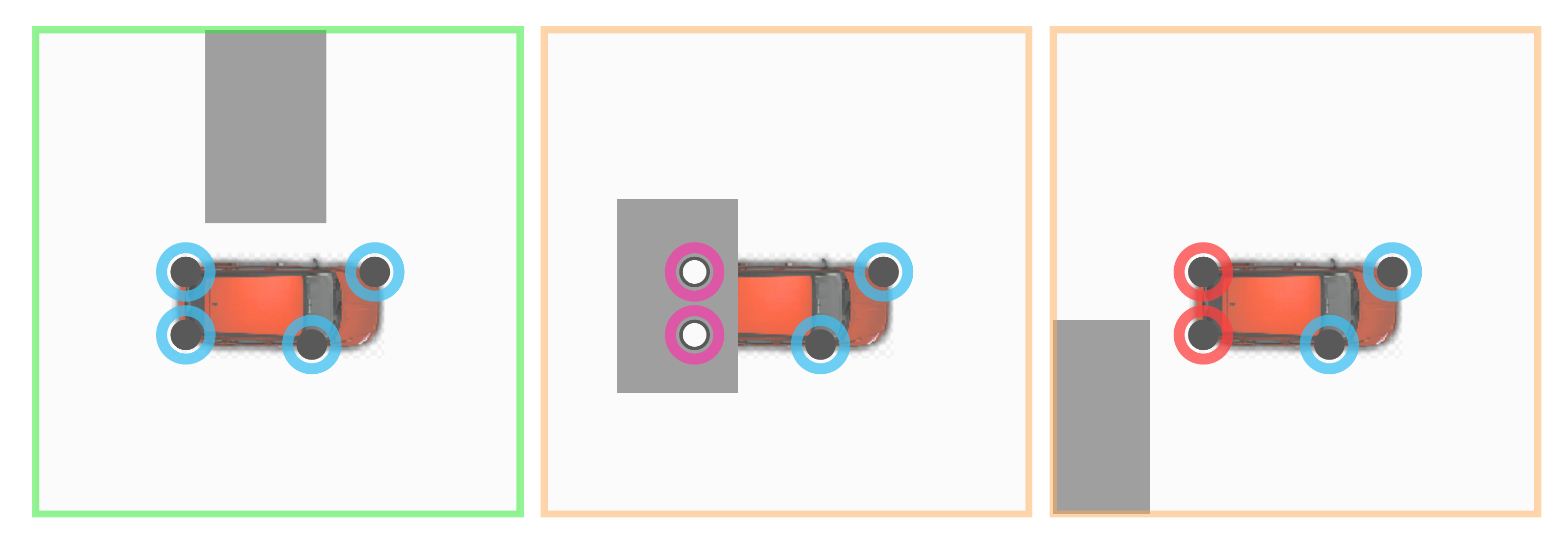}
  \label{fig:missing_points_lost}
}\\
\subfloat[]{
  \includegraphics[width=0.4\textwidth]{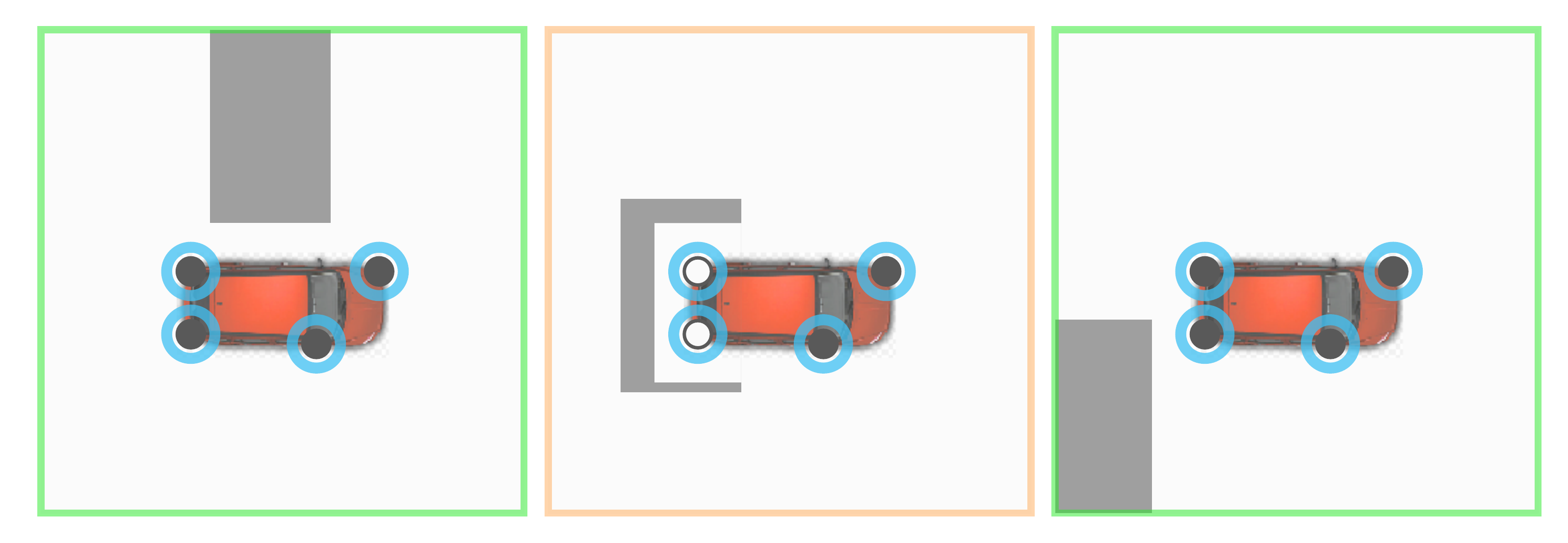}
  \label{fig:missing_points_recovered}
}
\caption{Three consecutive frames with partial occlusions in the middle frame.
(a) Keypoints lost without missing point reconstruction, and (b) keypoints
recovered with missing point reconstruction. \protect\tikz[baseline]
\protect\draw[color=cyan!65,semitransparent,line width=2pt] (0ex,0.5ex) circle
(0.75ex);-good points (flow success/recovered), \protect\tikz[baseline]
\protect\draw[color=magenta!65,semitransparent,line width=2pt] (0ex,0.5ex)
circle (0.75ex);-bad points (flow failure), \protect\tikz[baseline]
\protect\draw[color=red!65,opacity=0.75,line width=2pt] (0ex,0.5ex) circle
(0.75ex);-lost points, \protect\tikz[baseline]
\protect\filldraw[draw=darkgray!85!black!65,fill=darkgray!65,line width=1.5pt]
(0ex,0.5ex) circle (0.75ex);-visible points, \protect\tikz[baseline]
\protect\filldraw[draw=darkgray!65,fill=darkgray!2,line width=1.5pt] (0ex,0.5ex)
circle (0.75ex);-occluded points.}
\label{fig:missing_points}
\end{figure}

\subsubsection[From No Occlusion]{From No Occlusion ($\mathcal{O}^{old} = \mathcal{N}$)}
\label{subsubsec:from_no_occlusion}
In this situation we have $\vert {}^{good}\mathcal{F}^{old} \vert = n$ and
optical flow can be computed for feature points ${}^{good}\mathcal{F}^{old}$ to
obtain ${}^{good}\mathcal{F}^{new}$. The occlusion state identification in the
new frame ($\mathcal{O}^{new}$) can be carried out reliably as per
\eqref{eq:occulsion_inference}. Furthermore, by using the adjusted centroid,
missing keypoints ${}^{bad}\mathcal{F}^{new}$ can be back projected and
reconstructed as displayed in Fig.~\ref{fig:missing_points}. Occlusions cause
these points to be lost (Fig.~\ref{fig:missing_points_lost}), however (for cases
$1,2,3,5,6$) with reconstruction the missing points can be fully recovered
(Fig.~\ref{fig:missing_points_recovered}).

\begin{figure}
\centering
\includegraphics[width=0.4\textwidth]{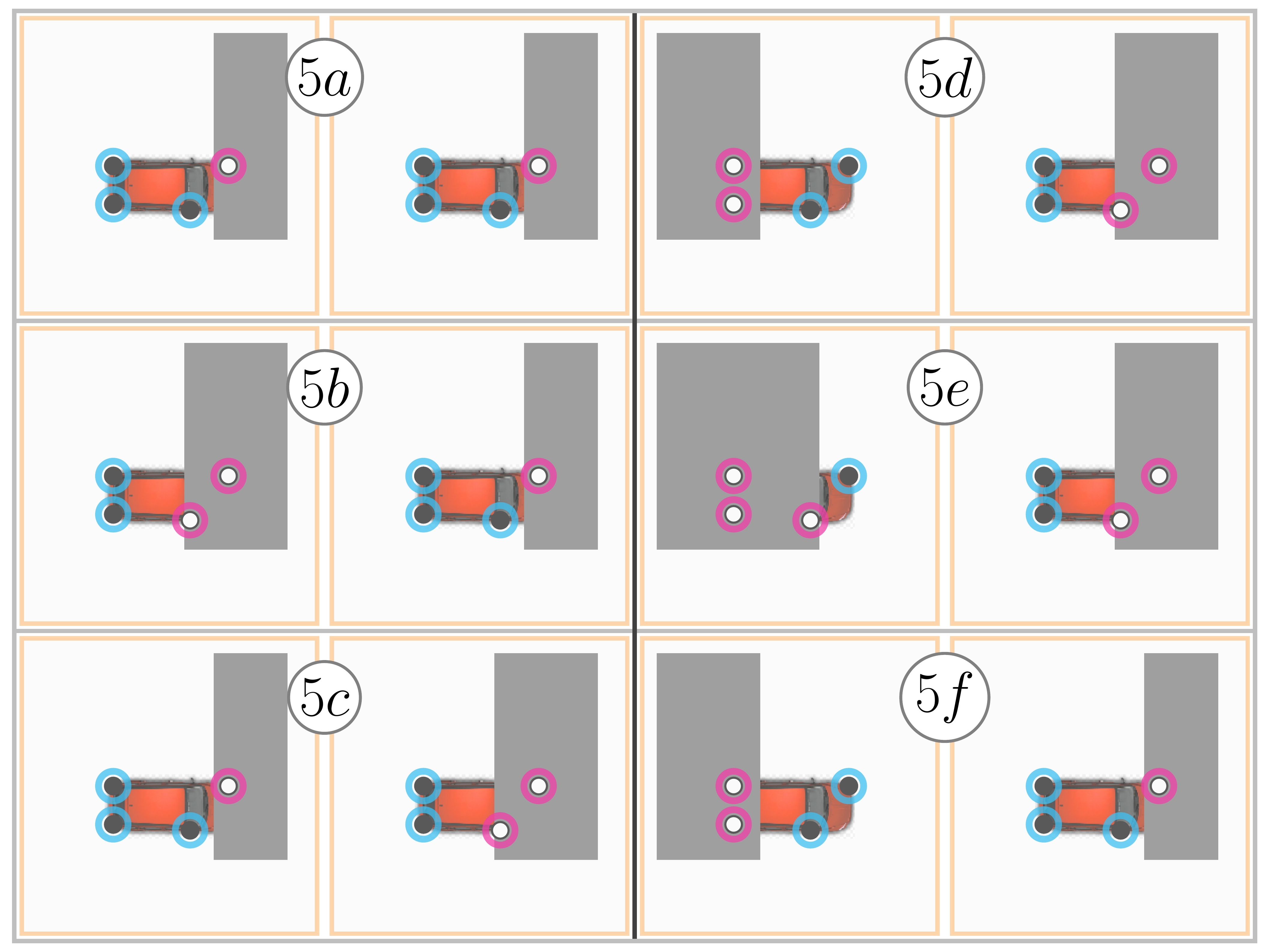}
\caption{Partial-partial occlusion state transition subcase instances. Subcases
($5a$-$5f$) represent equal, proper superset, proper subset, equal to
complement, proper subset of complement, and non-empty intersection,
respectively. \protect\tikz[baseline]
\protect\draw[color=cyan!65,semitransparent,line width=2pt] (0ex,0.5ex) circle
(0.75ex);-good points (flow success), \protect\tikz[baseline]
\protect\draw[color=magenta!65,semitransparent,line width=2pt] (0ex,0.5ex)
circle (0.75ex);-bad points (flow failure), \protect\tikz[baseline]
\protect\filldraw[draw=darkgray!85!black!65,fill=darkgray!65,line width=1.5pt]
(0ex,0.5ex) circle (0.75ex);-visible points, \protect\tikz[baseline]
\protect\filldraw[draw=darkgray!65,fill=darkgray!2,line width=1.5pt] (0ex,0.5ex)
circle (0.75ex);-occluded points. 
}\label{fig:subcases}
\end{figure}

\subsubsection[From Partial Occlusion]{From Partial Occlusion ($\mathcal{O}^{old} = \mathcal{P}$)}
\label{subsubsec:from_partial_occlusion}
Here we have $0 < \vert {}^{good}\mathcal{F}^{old} \vert \le n$ and optical flow
can be calculated at feature points ${}^{good}\mathcal{F}^{old}$. However, the
identification of occlusion state ($\mathcal{O}^{new}$) is not trivial and
\eqref{eq:occulsion_inference} cannot be reliably used. Therefore, we identify
the following six subcases (Fig.~\ref{fig:subcases}): (a)
${}^{good}\mathcal{F}^{new} = {}^{good}\mathcal{F}^{old}$, (b)
${}^{good}\mathcal{F}^{new} \supset {}^{good}\mathcal{F}^{old}$, (c)
${}^{good}\mathcal{F}^{new} \subset {}^{good}\mathcal{F}^{old}$, (d)
${}^{good}\mathcal{F}^{new} \cap {}^{good}\mathcal{F}^{old} = \varnothing$, (e)
${}^{good}\mathcal{F}^{new} \subset {}^{bad}\mathcal{F}^{old}$, and (f)
${}^{good}\mathcal{F}^{new} \cap {}^{good}\mathcal{F}^{old} \ne \varnothing$
(exclusive of other subcases).  Optical flow can be successfully determined in
subcases $5a, 5b, 5c, 5f$, or case $4$, while subcases $5d, 5e$, or case $6$
indicates failure. To perform robust tracking with occlusions we need to
accurately identify the occlusion states. In the partial-partial case, SIFT
feature matching or part-based template matching can be used. The subcase
treatments can be summarized as follows: $5a$ requires no further processing,
$5b$ necessitates redetection of the points, $5c$ involves missing point
reconstruction, $5f$ needs both redetection and missing point reconstruction,
$5d$ and $5e$ call for redetection using matching techniques.

\subsubsection[From Total Occlusion]{From Total Occlusion ($\mathcal{O}^{old} = \mathcal{T}$)}
\label{subsubsec:from_total_occlusion}
In this circumstance we have $\vert {}^{good}\mathcal{F}^{old} \vert = 0$, hence
optical flow computations cannot be made. To continue tracking for cases $7,8,9$
we can use techniques, such template or feature matching, and redetect
initial/updated feature points ${}^{good}\mathcal{F}^{0}$ in the new frame to
obtain ${}^{good}\mathcal{F}^{new}$. Occlusion state identification
($\mathcal{O}^{new}$) can then be decisively carried out as per
\eqref{eq:occulsion_inference}.

\subsection{Image Formation}
\label{subsec:image_formation}
Using optical flow, centroid adjustment, and occlusion handling mechanisms, we
can obtain noisy but accurate localization measurements of the vehicle
$\bm{x}_B$ from the image plane to the world inertial frame as follows. The
altitude ${}^{W}z_A(t)$ and camera calibration information such as field of view
($\vartheta=47^{\circ}$) and image sensor width ($\mathcal{W}=5\,mm$) are all
known. Therefore, with a pixel size of $6.25\,\mu m$ we can calculate the camera
focal length as 
\begin{equation}
  \label{eq:focal_length}
  \varphi = \frac{\mathcal{W}}{2}\left[ \tan\left(\frac{\vartheta}{2}\right) \right]^{-1}.
\end{equation}

Under orthographic projection of objects moving on a plane,
\eqref{eq:focal_length} can be used to obtain a mapping from image coordinates
$\bm{x}_B$ to world coordinates ${}^{W}\bm{x}_B$ by the geometric relation 
\begin{equation}
  \label{eq:altitude_relation}
  {}^{W}\bm{x}_B = \frac{{}^{W}z_A(t)}{\varphi}\bm{x}_B.
\end{equation}
Thus, \eqref{eq:altitude_relation} allows us to compute the measured kinematics
state vector $\left[{}^{W}\bm{x}_B^\top, {}^{W}\dot{\bm{x}}_B^\top\right]^\top$
of the vehicle in planar world coordinates. Finally, $\left[{}^{W}{x}_B,
{}^{W}{y}_B\right]^\top$ is converted to polar form ($r,\theta$).

\section{Simulations}  
\label{sec:simulations}
In this section, we demonstrate the behavior of our vision-based guidance
framework using the following (unknown to the UAS) vehicle trajectory variants:
lane changing and squircle following. All simulations were performed on a
Windows 10 machine with an Intel Core i7-8700 CPU and 32 GB RAM. For every
simulation, we make use of the true values of each quantity alongside their
measured and/or estimated values to generate the data plots.

\subsubsection{Lane Changing Trajectory}
\label{subsubsec:case_1}
\begin{figure}
\centering
\subfloat[]{
  \includegraphics[width=\simplotwidth]{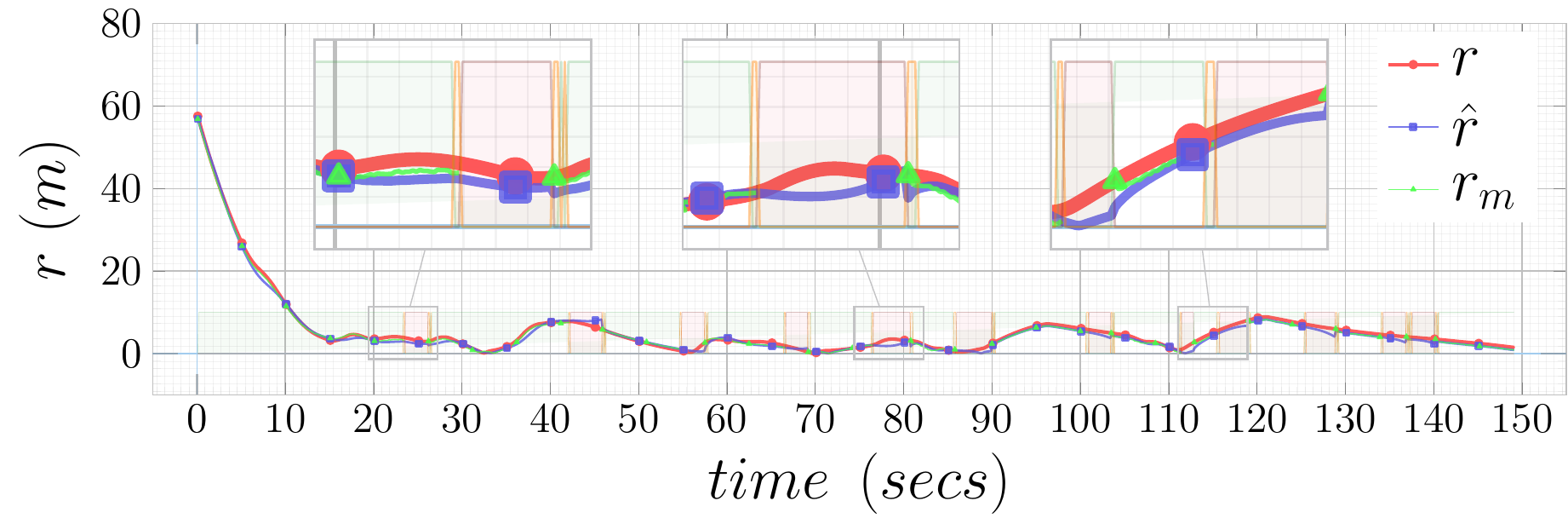}
}\\
\subfloat[]{
  \includegraphics[width=\simplotwidth]{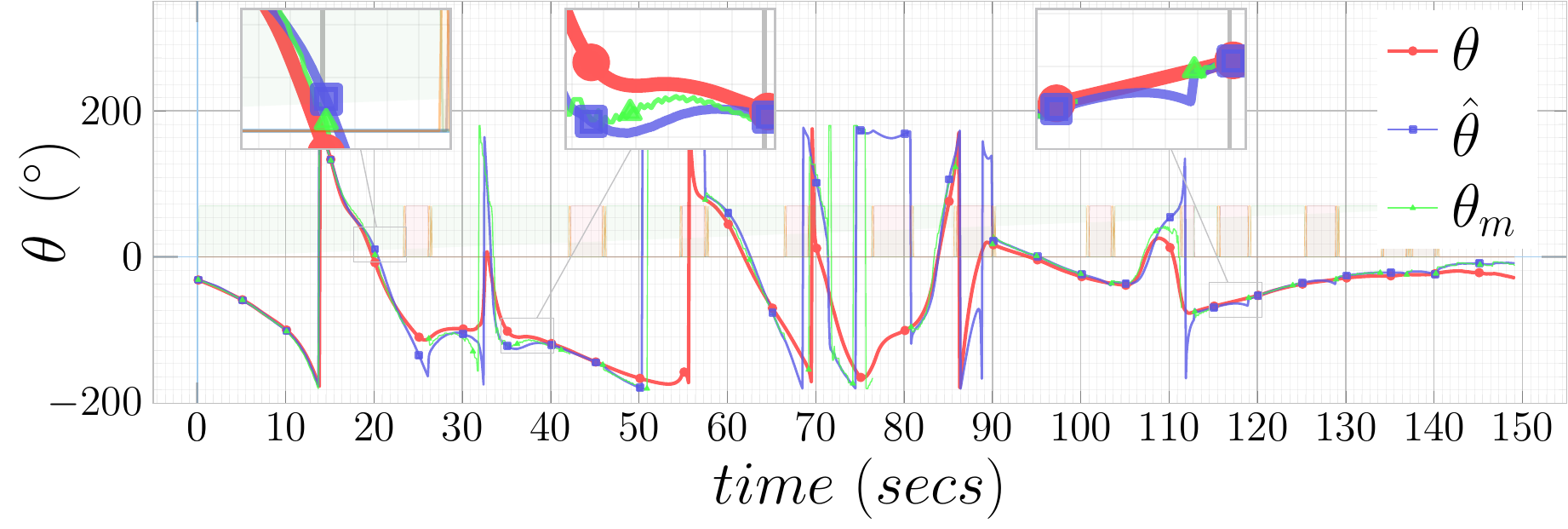}
}\\
\subfloat[]{
  \includegraphics[width=\simplotwidth]{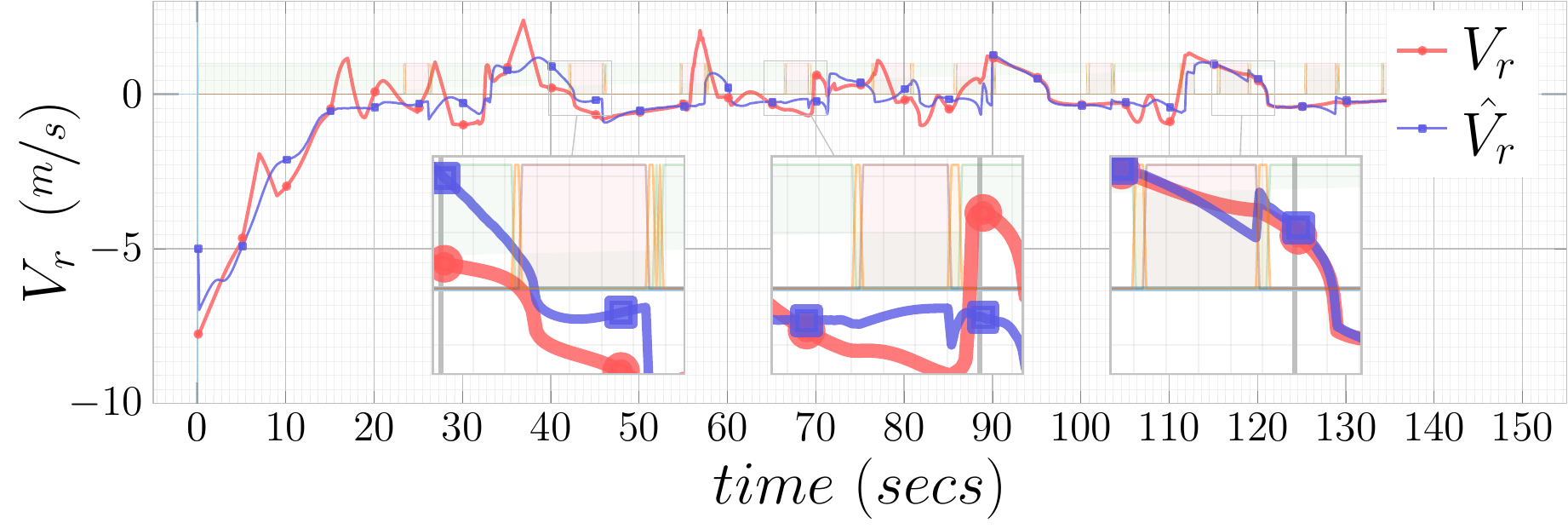}
}\\ 
\subfloat[]{
  \includegraphics[width=\simplotwidth]{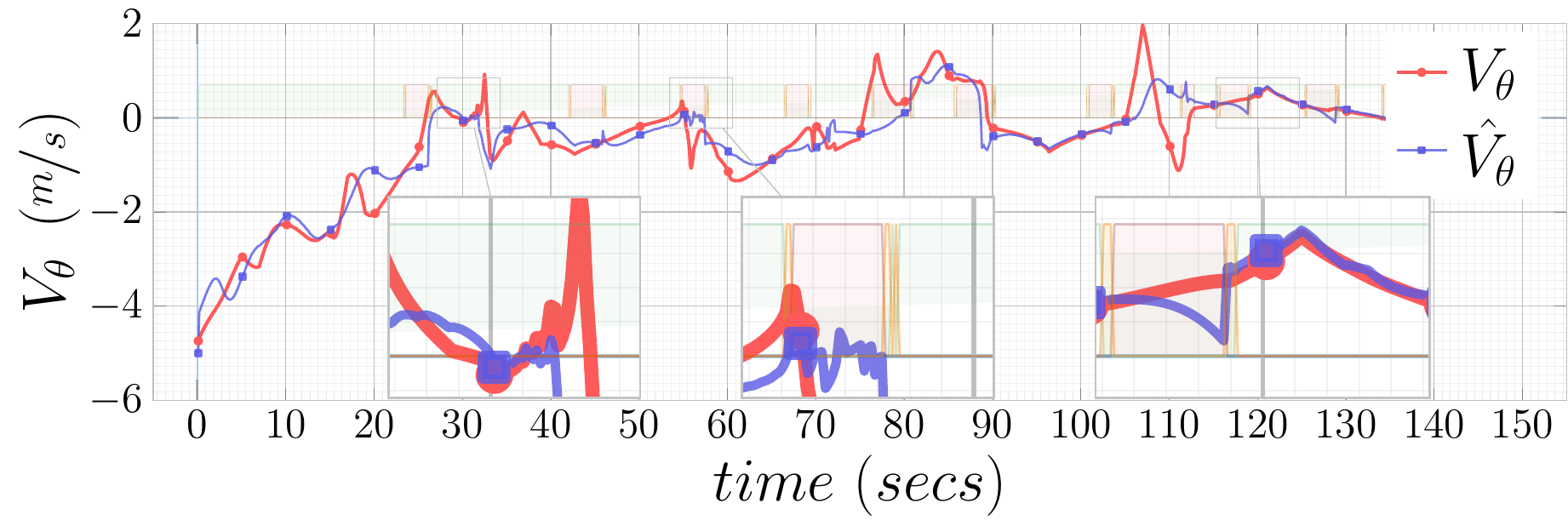}
 }\\
\subfloat[]{
  \includegraphics[width=\simplotwidth]{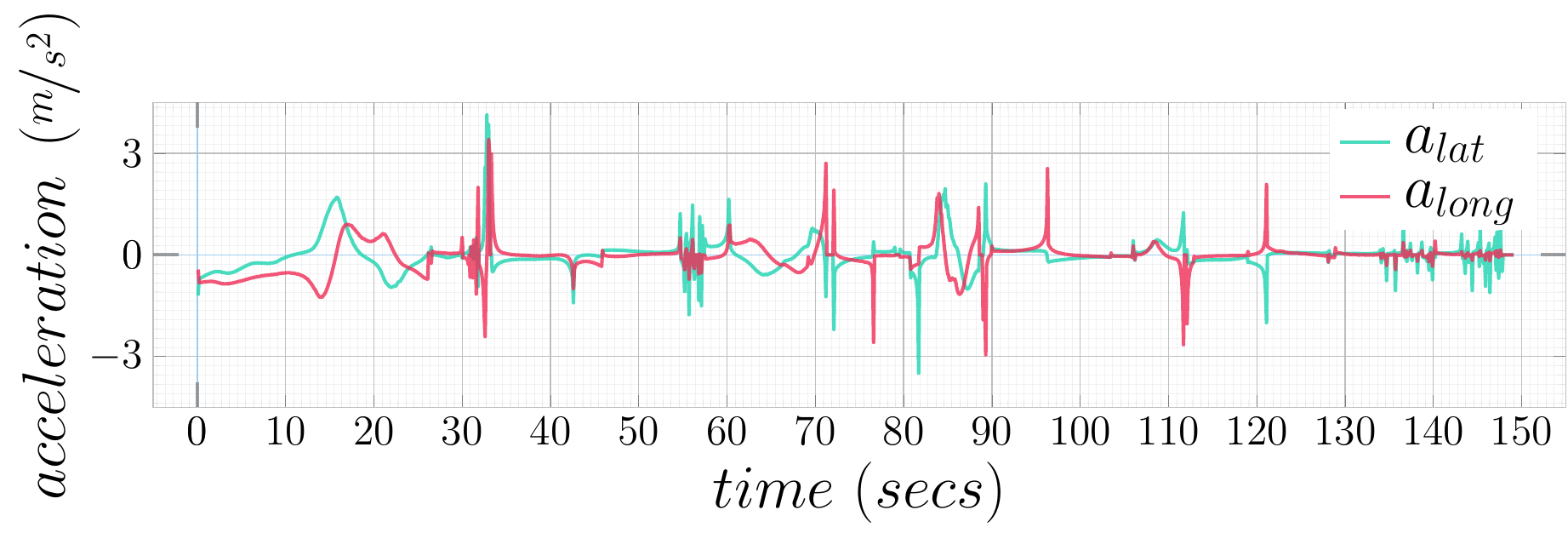}
}
\caption{Lane changing trajectory: (a)-(d) true, estimated, and measured states
$r$, $\theta$, $V_r$, and $V_{\theta}$; (e) acceleration commands $a_{lat}$ and
$a_{long}$. No (\protect\tikz[baseline] \protect\filldraw[color=BLOOD_RED,
fill=BLOOD_RED!20, line width=0.5pt, opacity=0.5] (0ex,-0.5ex) rectangle
(3ex,1.5ex);), partial (\protect\tikz[baseline]
\protect\filldraw[color=YELLOW_ORANGE, fill=YELLOW_ORANGE!20, line width=0.5pt,
opacity=0.5] (0ex,-0.5ex) rectangle (3ex,1.5ex);), and total
(\protect\tikz[baseline] \protect\filldraw[color=GREEN_PANTONE,
fill=GREEN_PANTONE!20, line width=0.5pt, opacity=0.5] (0ex,-0.5ex) rectangle
(3ex,1.5ex);) occlusion states are indicated along the zero line.}
\label{fig:lane_change_1}
\end{figure}

\begin{figure}
\centering
\subfloat[]{
  \includegraphics[width=\simplotwidth]{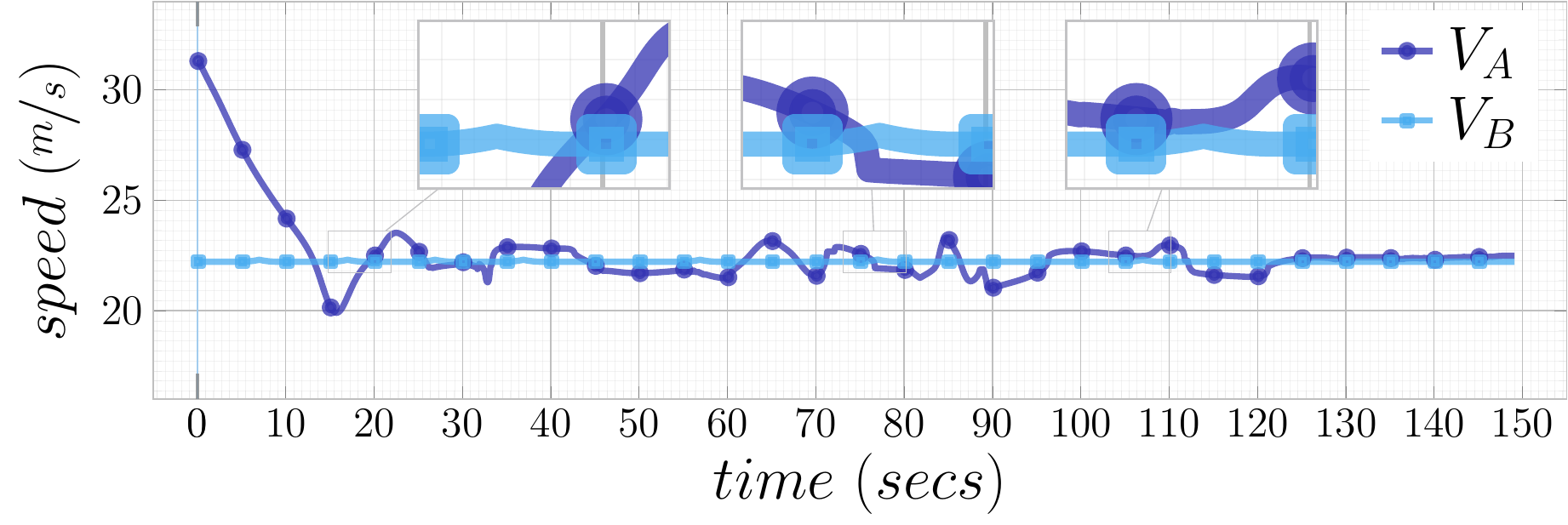}
}\\
\subfloat[]{
  \includegraphics[width=\simplotwidth]{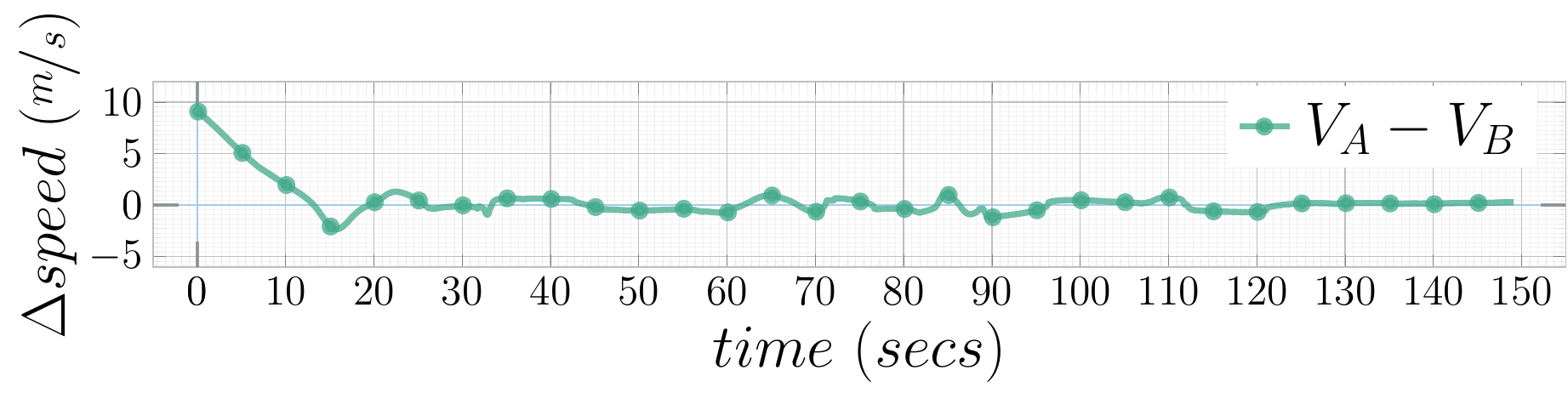}
}\\
\subfloat[]{
  \includegraphics[width=\simplotwidth]{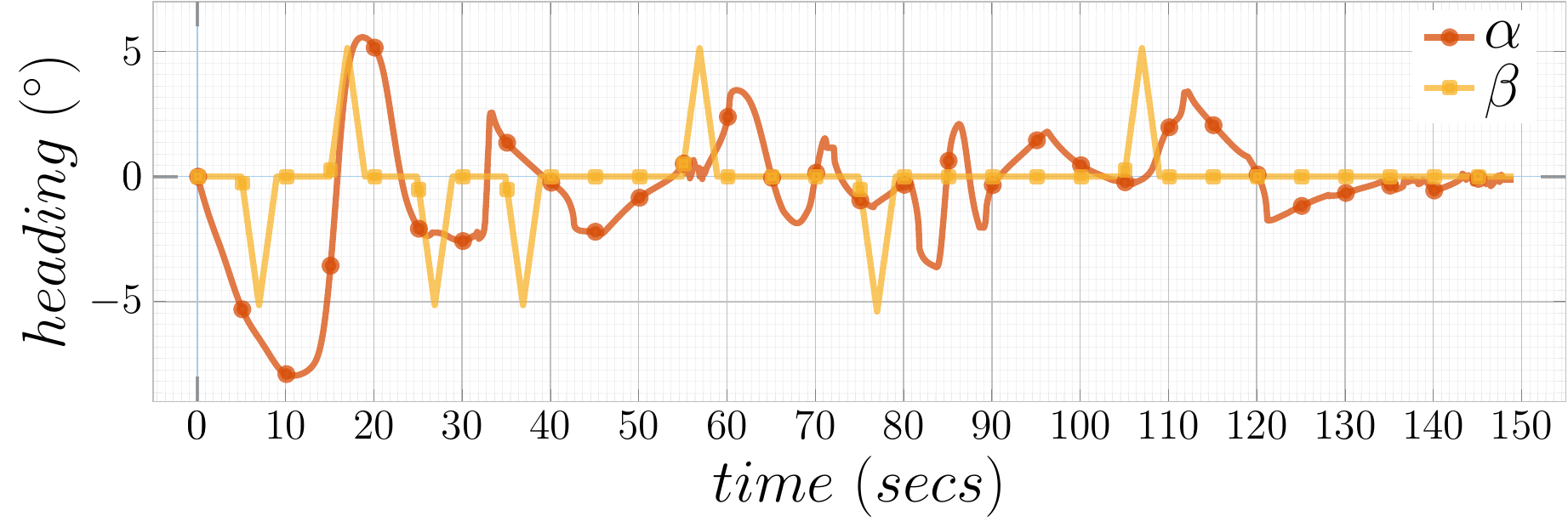}
}\\ 
\subfloat[]{
  \includegraphics[width=\simplotwidth]{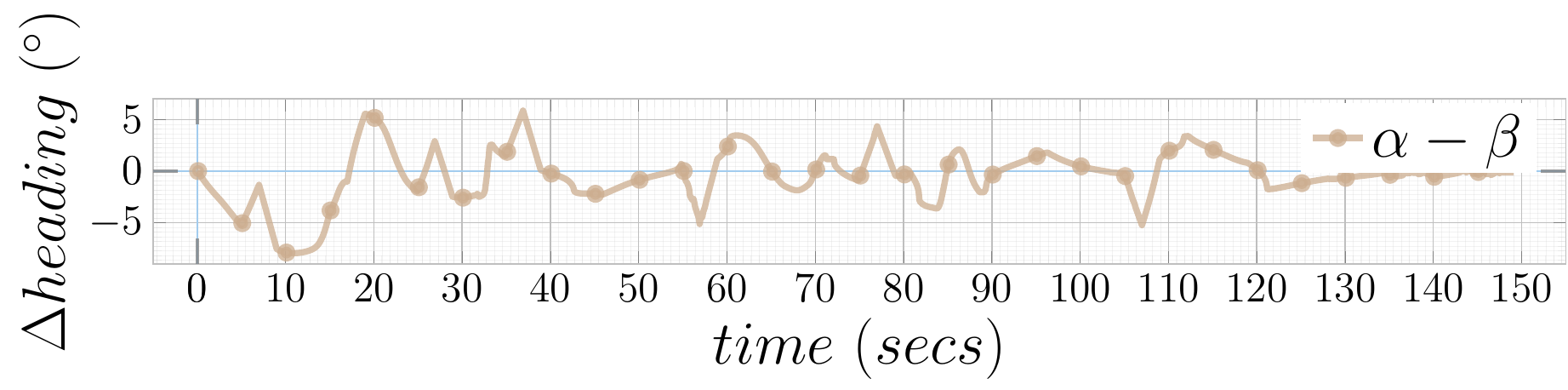}
}\\ 
\subfloat[]{
  \includegraphics[width=\simplotwidth]{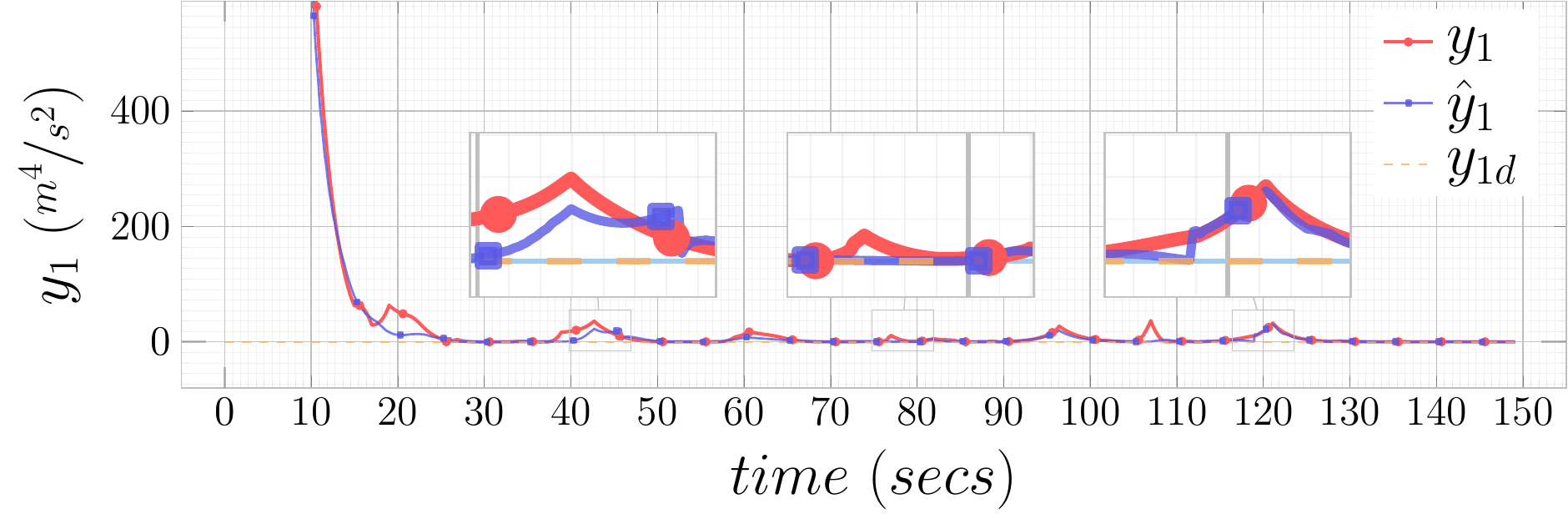}
}\\
\subfloat[]{
  \includegraphics[width=\simplotwidth]{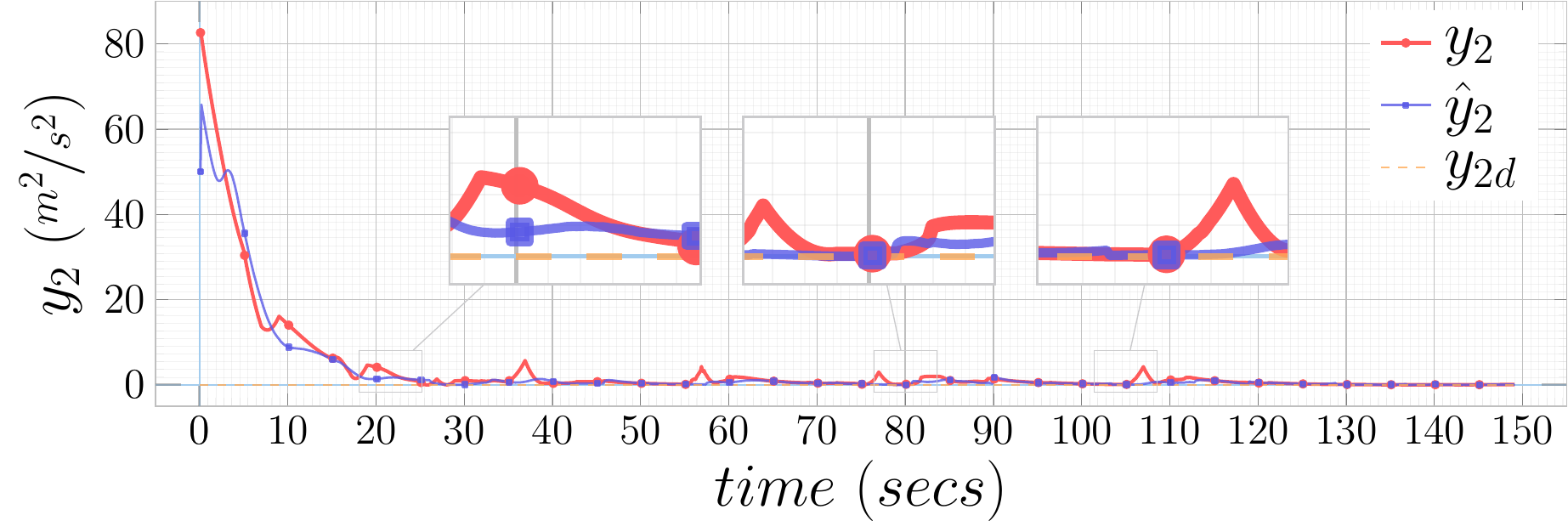}
}
\caption{Lane changing trajectory: (a) speed of the UAS and vehicle and (b)
their differences; (c) heading of the UAS and vehicle and (b) their differences;
(e)-(f) true and estimated objective functions $y_1$ and $y_2$.}
\label{fig:lane_change_2}
\end{figure}

\begin{figure}
\centering
\subfloat[]{
  \includegraphics[width=\simplotwidth]{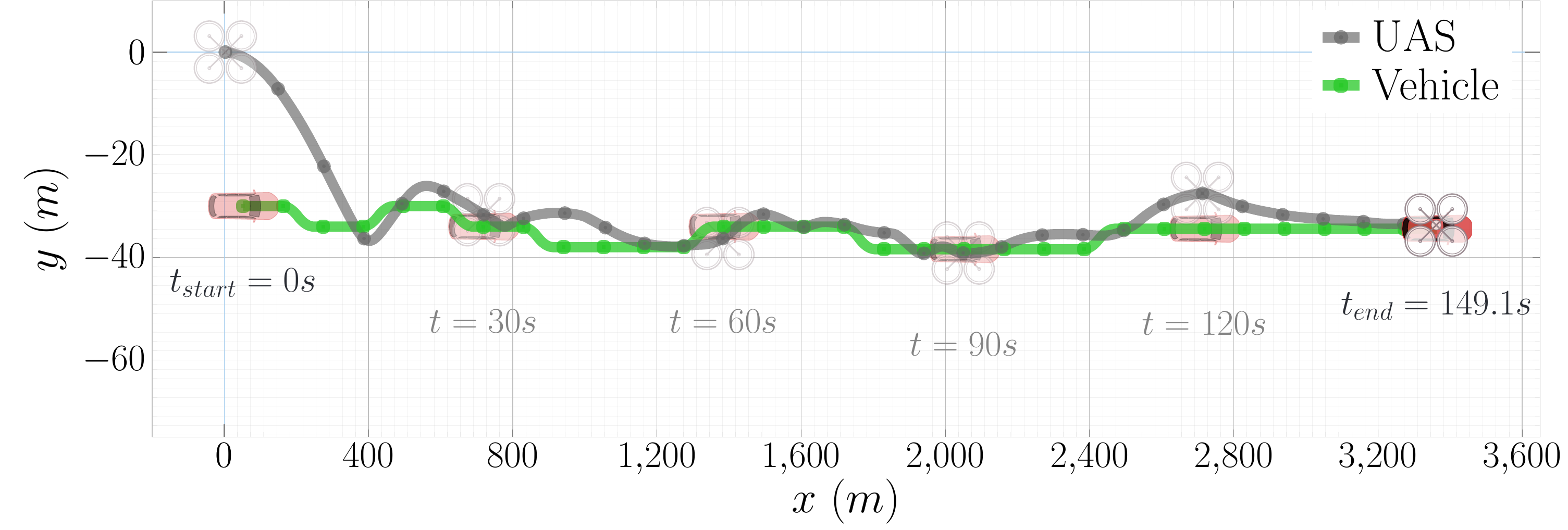}
}\\
\subfloat[]{
  \includegraphics[width=\simplotwidth]{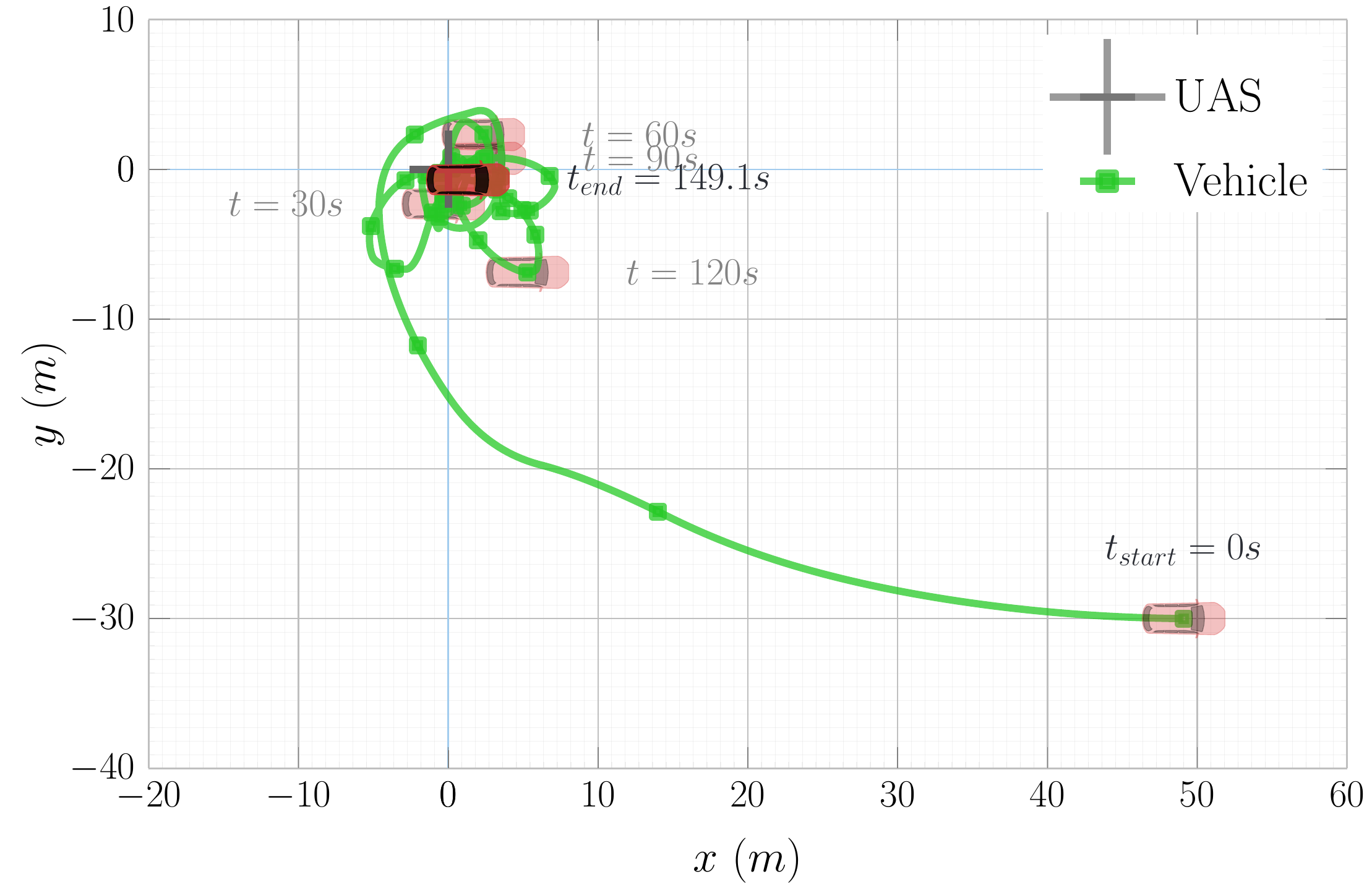}
}
\caption{Lane changing trajectory: UAS and vehicle trajectories in world (a) and
camera (b) reference frames.}
\label{fig:lane_change_3}
\end{figure}

We simulate the vehicle to move in the east direction while performing lane
changing maneuvers of approximately $4\,m$ at arbitrary time instances with
occlusions. The UAS starts at position ${}^{W}{\bm{x}_{A}}_0 = [0.0, 0.0,
150.0]^\top (m)$ while the vehicle begins at ${}^{W}{\bm{x}_{B}}_0 = [50.0,
-30.0, 0.0]^\top (m)$ with speeds ${}^{W}{V_{A}}_0 = 31.31\ (\nicefrac{m}{s})$
($70\,mph$) and ${}^{W}{V_{B}}_0 = 22.22\ (\nicefrac{m}{s})$ ($50\,mph$), and
headings ${}^{W}{\alpha}_{0}=0.0^{\circ}$ and ${}^{W}{\beta}_{0}=0.0^{\circ}$,
respectively. The simulation was run for approximately $149\,secs$.
Fig.~\ref{fig:lane_change_1} depicts the kinematic state information collected
over time. Under total occlusions, measurements disappear and state estimations
are utilized to perform guidance. It can be observed in the data plots that
estimations may veer away from the true values if an occlusion persists for a
long time. In the simulations, the occurrence of occlusions is accompanied by an
increasing covariance ($\bm{R}_B$) in the measurement noise ($\bm{v}_B$) in
\eqref{eq:motion_model} and by employing previous estimates as current
measurements in the acceleration model. In Fig.~\ref{fig:lane_change_2}, the
speed and heading profiles of the UAS and vehicle along with their deltas are
provided. The trajectories of the UAS and vehicle in world and camera reference
frames are shown in Fig.~\ref{fig:lane_change_3}. Visual tracking was
successfully performed using our occlusion handling and rendezvous cone-based
guidance scheme.

\subsubsection{Squircle Following Trajectory}
\label{subsubsec:case_2}
\begin{figure}
\centering
\subfloat[]{
  \includegraphics[width=\simplotwidth]{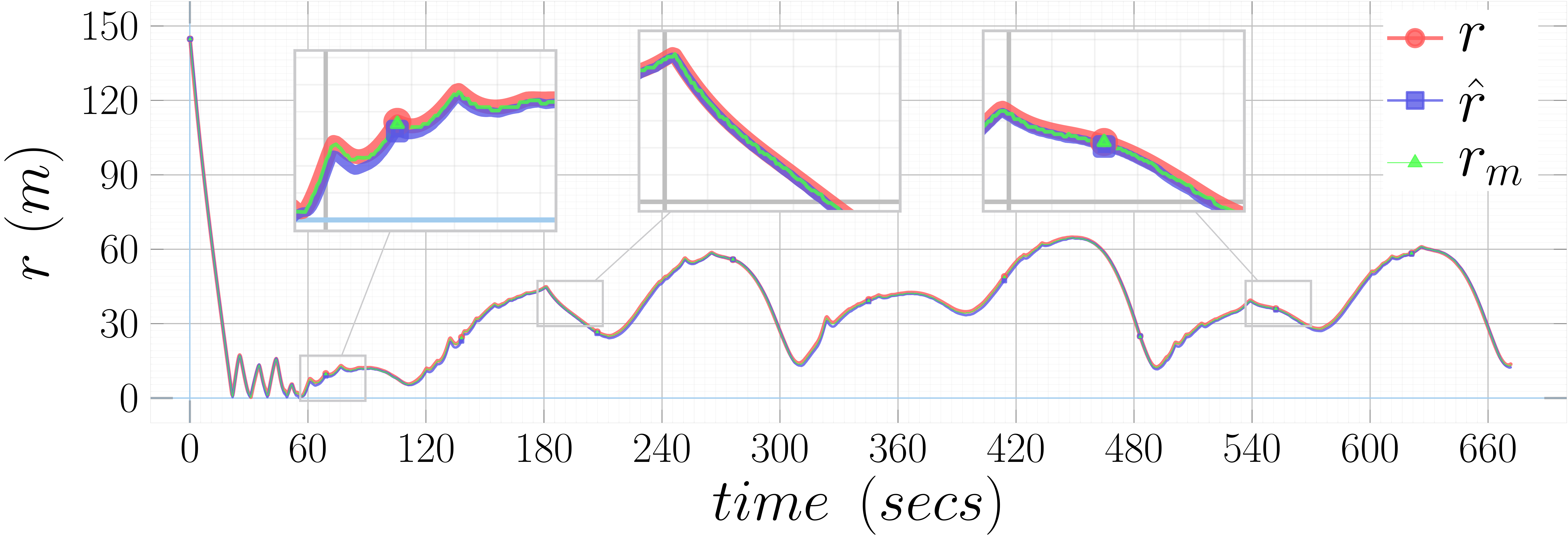}
}\\
\subfloat[]{
  \includegraphics[width=\simplotwidth]{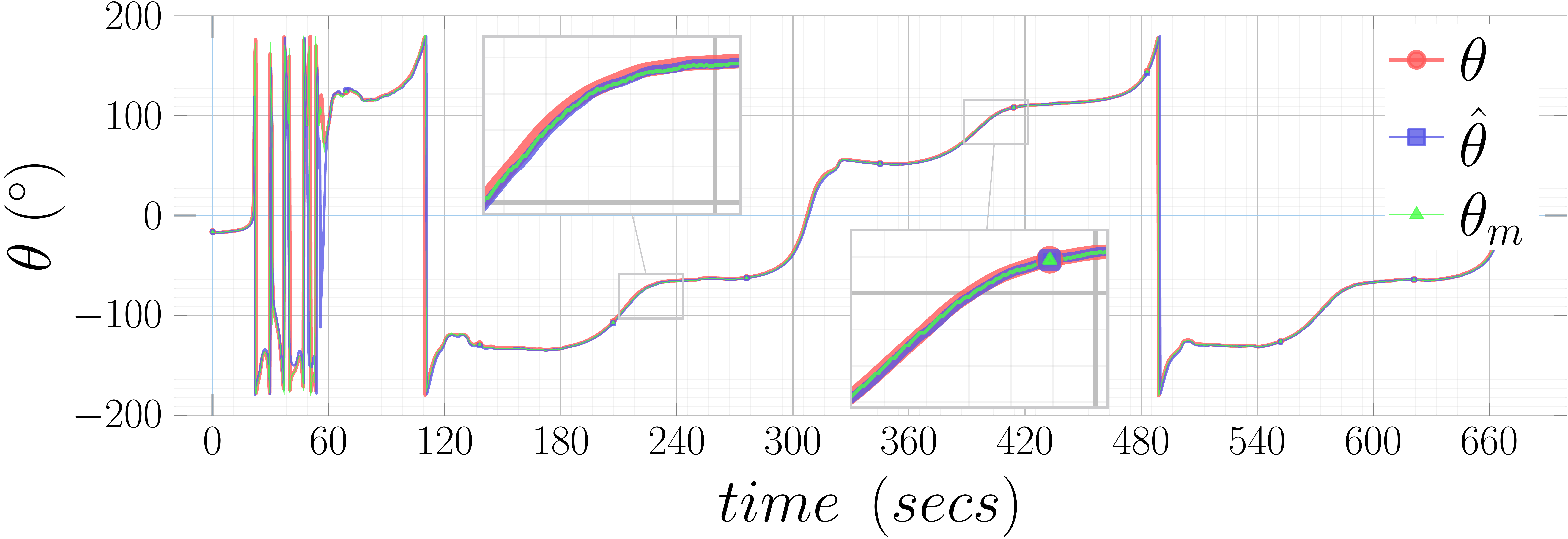}
}\\
\subfloat[]{
  \includegraphics[width=\simplotwidth]{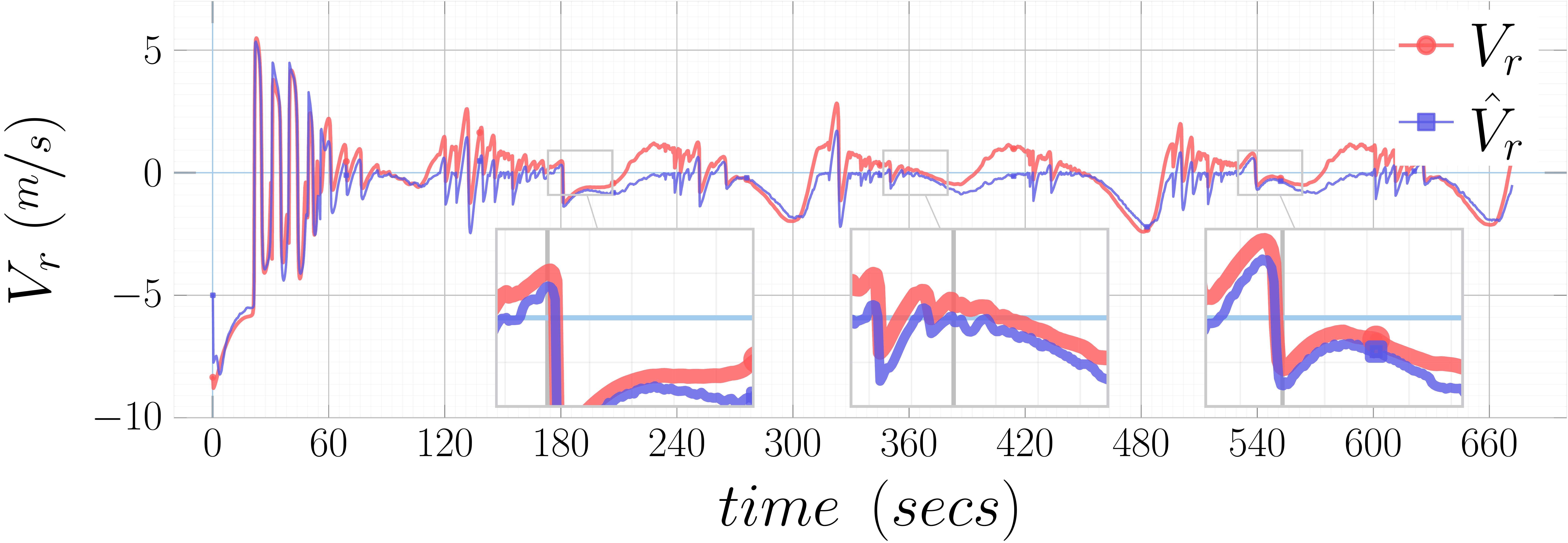}
}\\ 
\subfloat[]{
  \includegraphics[width=\simplotwidth]{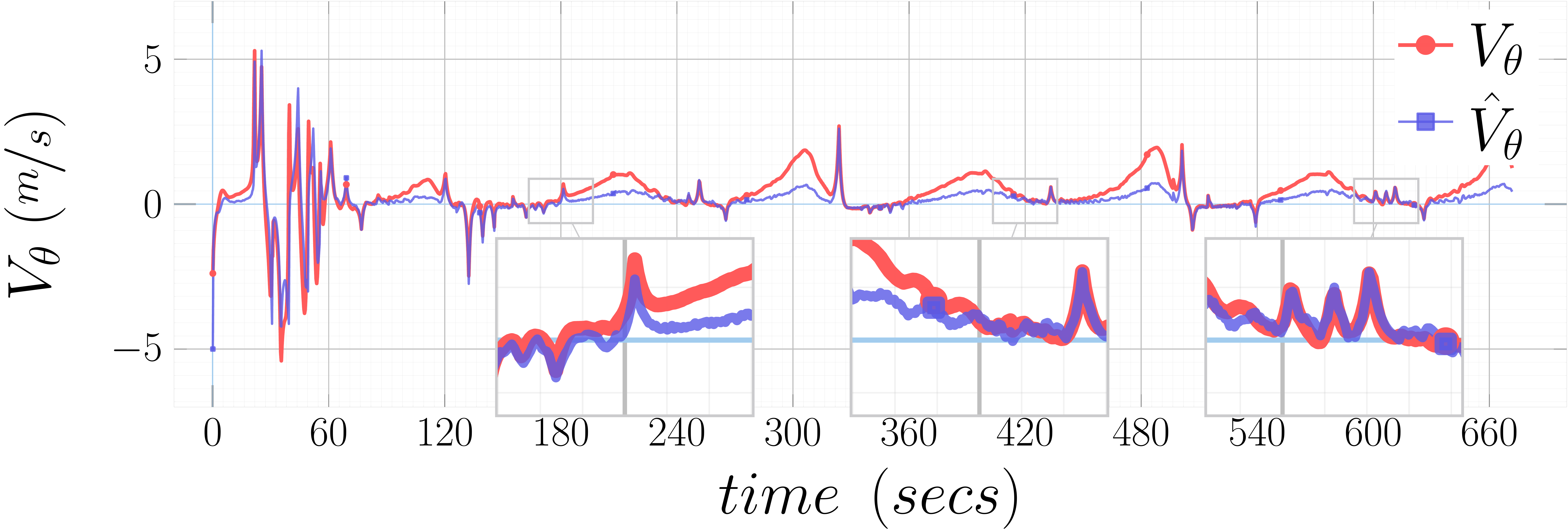}
}\\
\subfloat[]{
  \includegraphics[width=\simplotwidth]{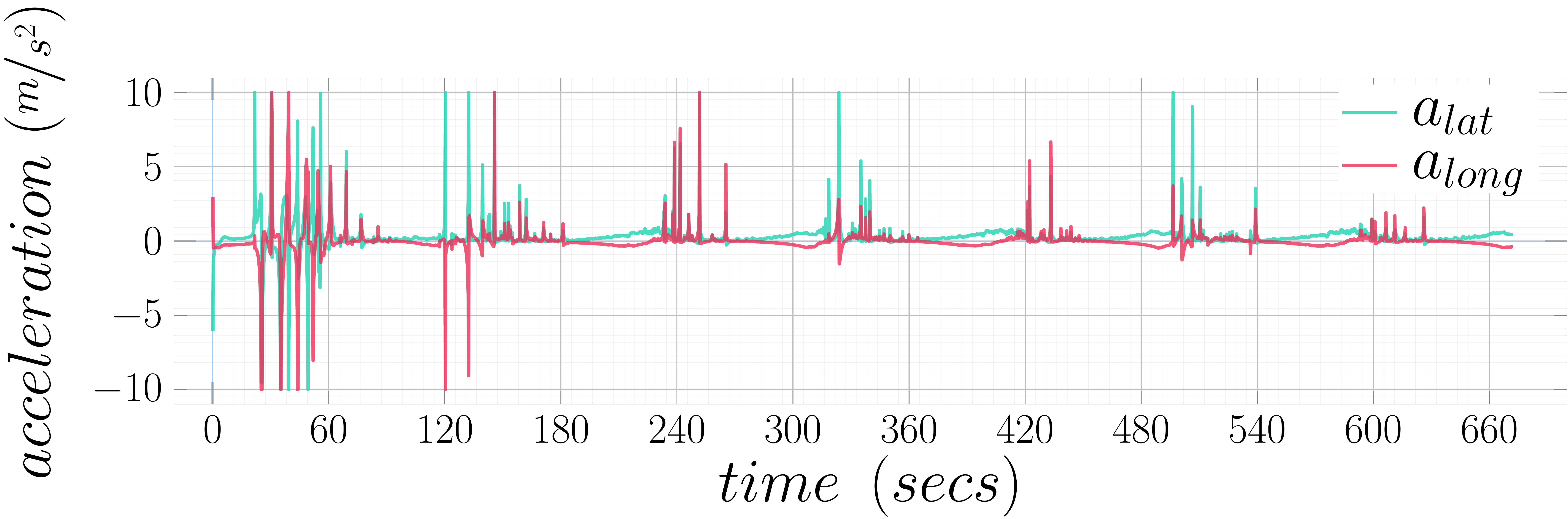}
}
\caption{Squircle following trajectory: (a)-(d) true, estimated, and measured
states $r$, $\theta$, $V_r$, and $V_{\theta}$; (e) acceleration commands
$a_{lat}$ and $a_{long}$.} 
\label{fig:squircle_1}
\end{figure}

\begin{figure}
\centering
\subfloat[]{
  \includegraphics[width=\simplotwidth]{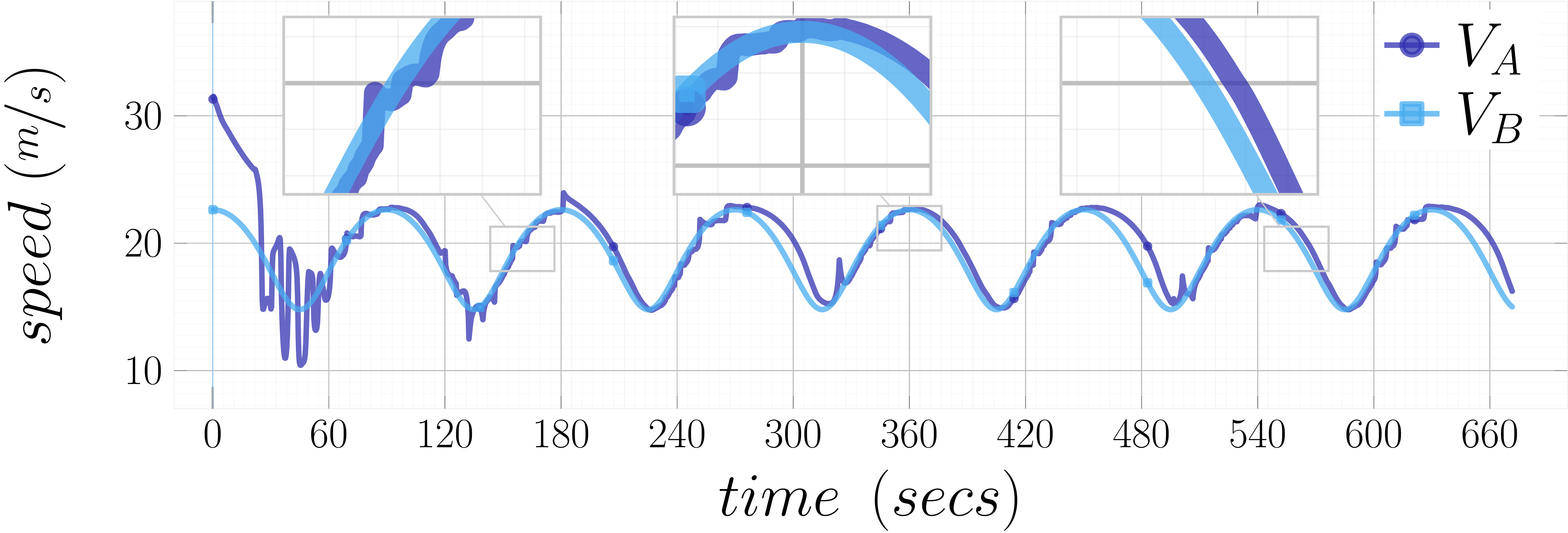}
}\\
\subfloat[]{
  \includegraphics[width=\simplotwidth]{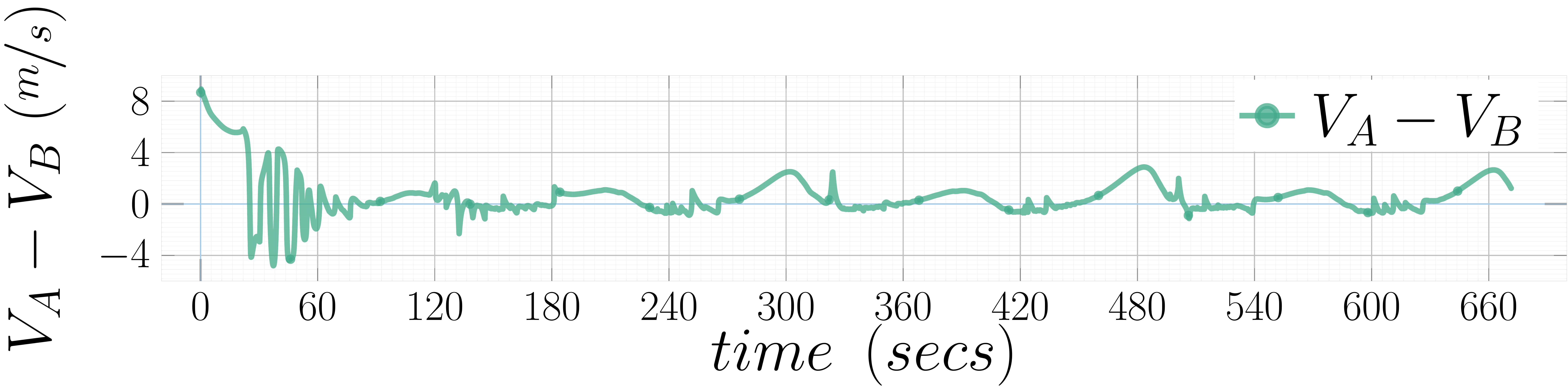}
}\\
\subfloat[]{
  \includegraphics[width=\simplotwidth]{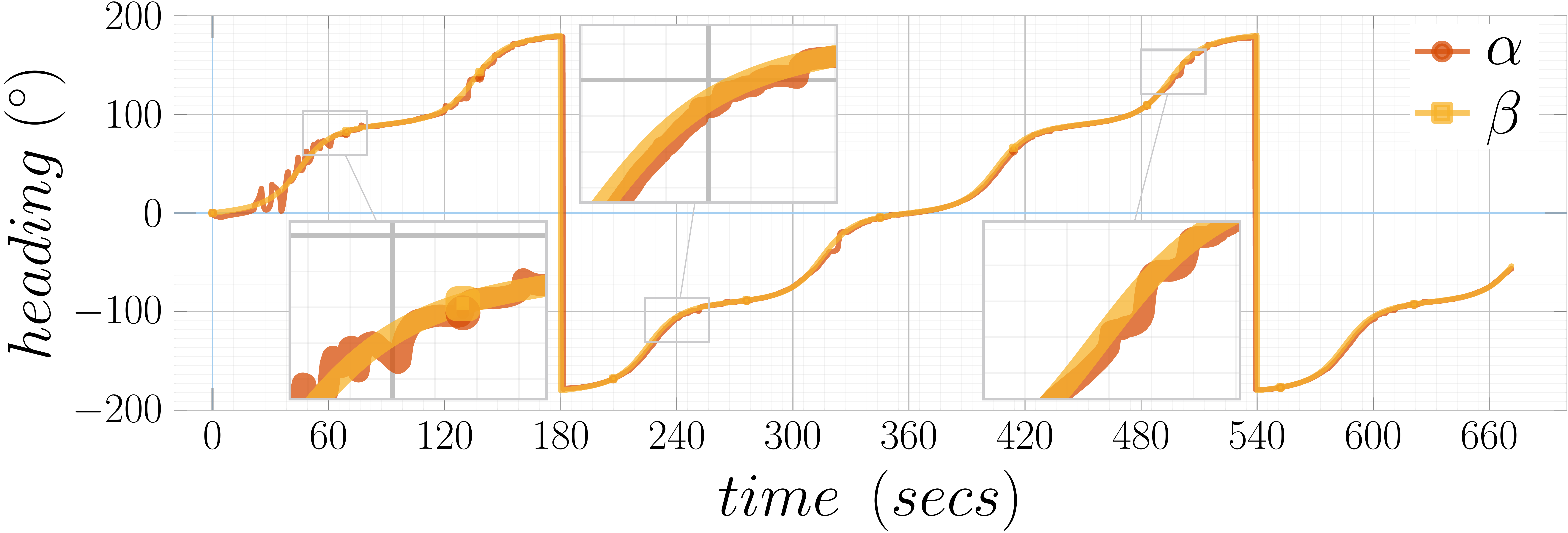}
}\\ 
\subfloat[]{
  \includegraphics[width=\simplotwidth]{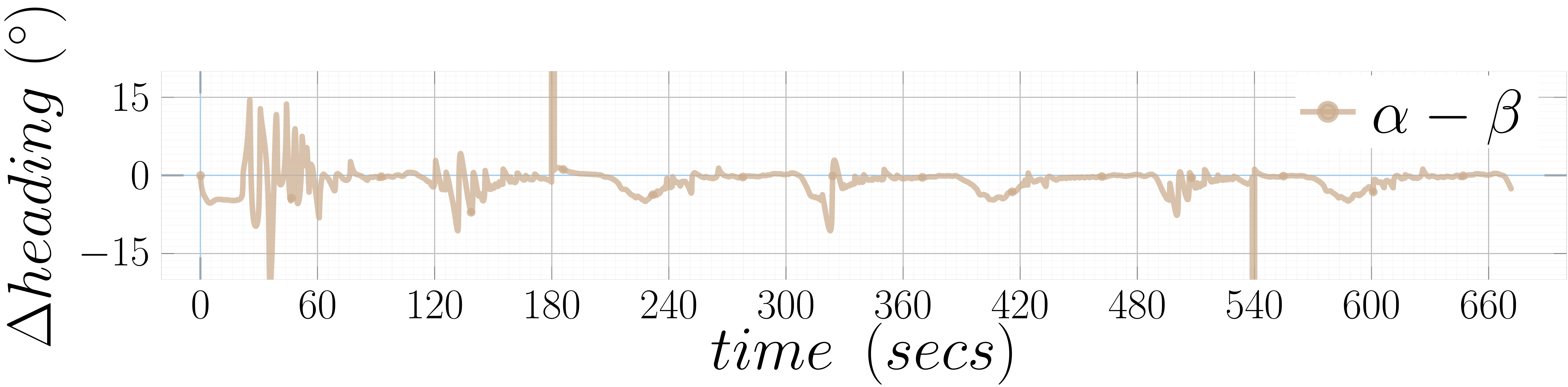}
}\\ 
\subfloat[]{
  \includegraphics[width=\simplotwidth]{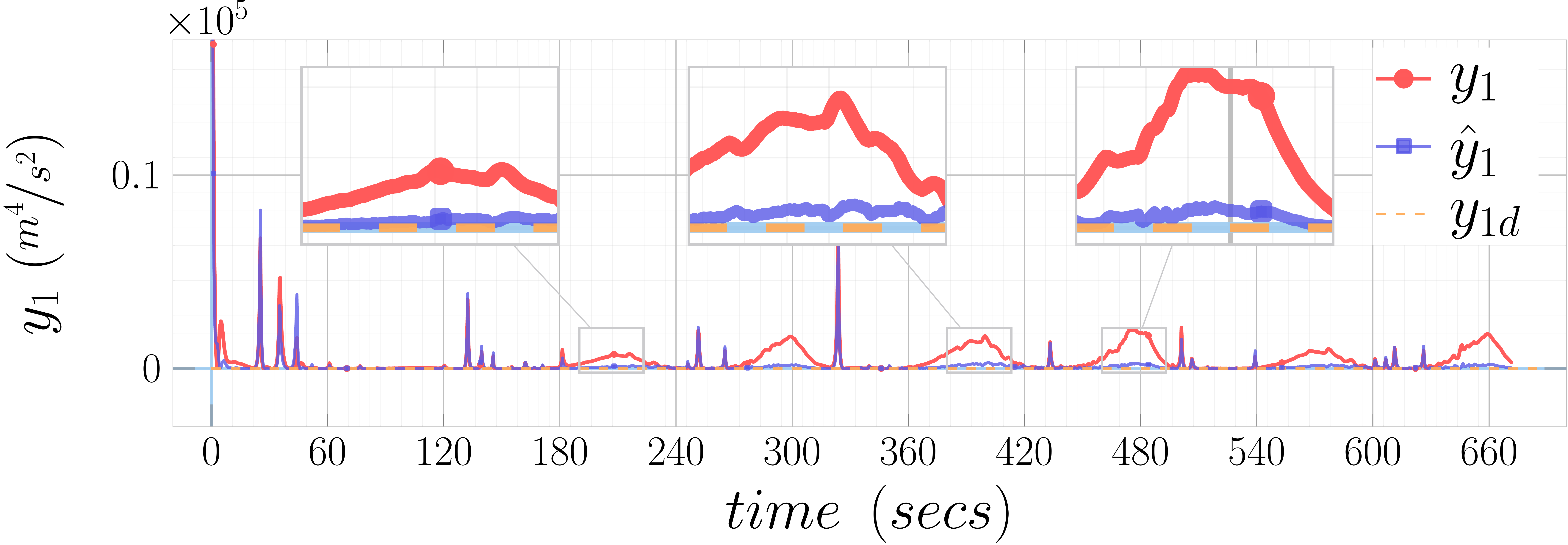}
}\\
\subfloat[]{
  \includegraphics[width=\simplotwidth]{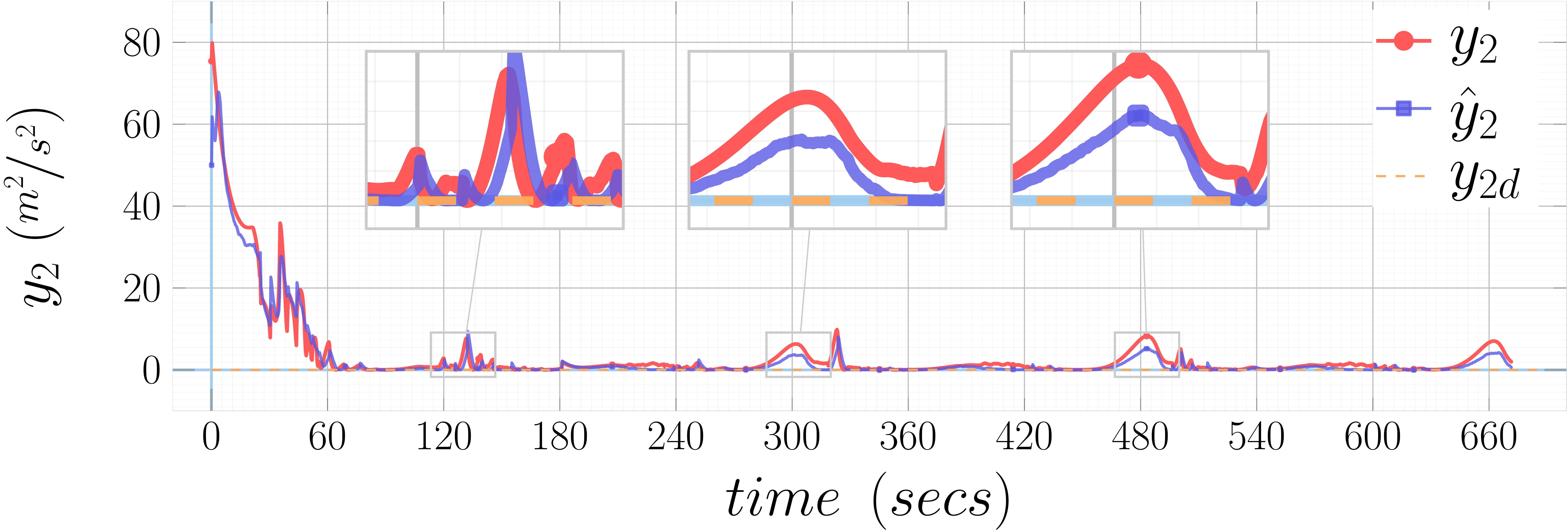}
}
\caption{Squircle following trajectory: (a) speed of the UAS and vehicle and (b)
their differences; (c) heading of the UAS and vehicle and (d) their differences;
(e)-(f) true and estimated objective functions $y_1$ and $y_2$.}
\label{fig:squircle_2}
\end{figure}

\begin{figure}
\centering
\subfloat[]{
  \includegraphics[width=\simplotwidth]{./case_2/plot_traj_world.pdf}
}\\
\subfloat[]{
  \includegraphics[width=\simplotwidth]{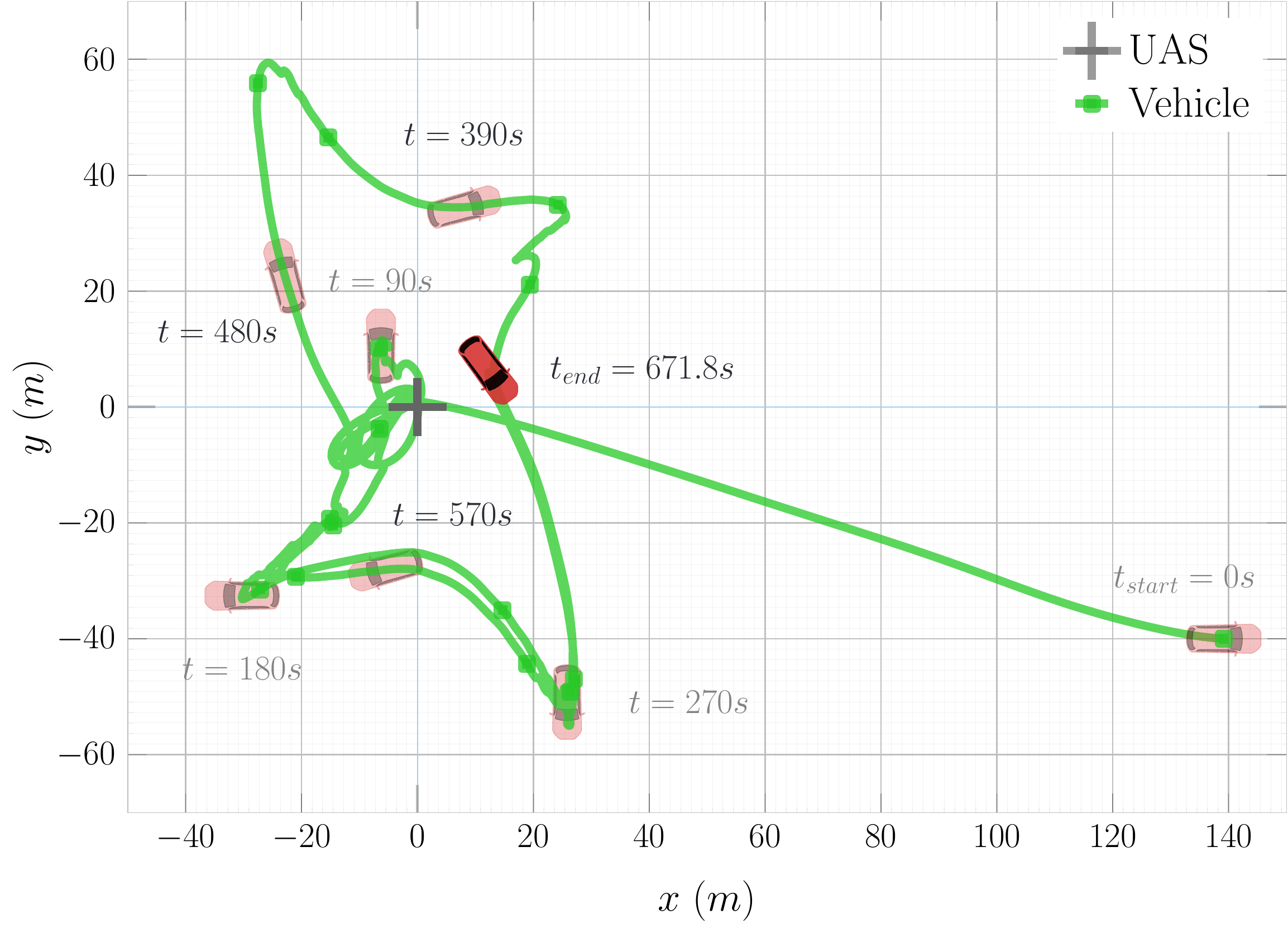}
}
\caption{Squircle following trajectory: UAS and vehicle trajectories in world
(a) and camera (b) frames of reference.}
\label{fig:squircle_3}
\end{figure}

We simulate the vehicle to proceed along a parameterized squircular trajectory
such that it traverses the closed curve with a period of $360\,secs$. The UAS
begins at position ${}^{W}{\bm{x}_{A}}_0 = [0.0, 0.0,350.0]^\top (m)$ with speed
${}^{W}{V_{A}}_0 = 31.31\ (\nicefrac{m}{s})$ ($70\,mph$) while the vehicle
starts at ${}^{W}{\bm{x}_{B}}_0 = [0.0, -20.0, 0.0]^\top (m)$. Both the UAS and
vehicle have an initial heading of ${}^{W}{\alpha}_{0}=0.0^{\circ}$ and
${}^{W}{\beta}_{0}=0.0^{\circ}$. The vehicle speed ${}^{W}{V_{B}}$ decreases
around the corners to approximately $14.79\ (\nicefrac{m}{s})$ ($33.08\,mph$)
while it reaches nearly $22.63\ (\nicefrac{m}{s})$ ($50.62\,mph$) along the
straight ways. The simulation was run for about $671\,secs$. In
Fig.~\ref{fig:squircle_1}, the true, estimated, and measured kinematics states
along with the lateral and longitudinal accelerations are shown. The speed,
heading, and deltas are visualized in Fig.~\ref{fig:squircle_2} along with the
true and estimated objective functions $y_1$ and $y_2$.
Fig.~\ref{fig:squircle_3} depicts the trajectories of the UAS and vehicle in
world and camera frames of reference. Our vision-based guidance system was able
to successfully allow the UAS to chase the vehicle while matching its speed and
heading at non-constant velocities.

\section{Conclusion and Future Work}
\label{sec:conclusion}
In conclusion, we have presented our solution to a problem wherein a UAS must
track a ground-based vehicle moving at erratic velocities through occlusions
using only visual information from a single camera. A rendezvous cone guidance
scheme drives our objective functions to the desired values resulting in the
generation of acceleration commands that lead the UAS into the cone and towards
the vehicle. We perform visual tracking of the vehicle's centroid using feature
detection and optical flow along with a diagnosis and analysis of occlusion
events. A publicly available simulator that provides an environment to perform
these vision-based guidance simulations has been developed. In the future, we
plan to extend the guidance strategy to 3D, incorporate multi-vehicle tracking,
and improve the fidelity of the simulator. 

%
%

\section*{Acknowledgments}
This material is based in part upon work supported by the National Science
Foundation through grant \#IIS-1851817 and a University of Texas at Arlington
Research Enhancement Program grant \#270079. 

\bibliographystyle{IEEEtran}
\bibliography{IEEEabrv,vision-based_guidance_for_tracking_dynamic_objects}   
\end{document}